%% file: main.tex
\documentclass[10pt]{article} 
\usepackage[preprint]{rlj}           

\usepackage{amssymb}
\usepackage{mathtools}
\usepackage{mathrsfs}
\usepackage{graphicx}
\usepackage{subcaption}
\usepackage[space]{grffile}
\usepackage{url}
\usepackage{booktabs}
\usepackage{enumitem}

\title{Evaluating Feature Dependent Noise in Preference-based Reinforcement Learning}

\setrunningtitle{Feature-Dependent Noise in Preference-based RL}

\author{Yuxuan Li\textsuperscript{1}, Harshith Reddy Kethireddy\textsuperscript{2}, Srijita Das\textsuperscript{2}}

\emails{yuxuan.li1@uwaterloo.ca, kharshi@umich.edu, sridas@umich.edu}

\affiliations{
$^{1}$\textbf{University of Waterloo, Waterloo, Canada}\\
$^{2}$\textbf{University of Michigan - Dearborn, Dearborn, United States}
}

\contribution{
We formalise feature-dependent noise in context of Preference-based Reinforcement Learning providing a foundation for structured, feature-correlated uncertainty in preference data.
}{
Recent noise-robust PbRL algorithms, such as RIME~\citep{Cheng2024RIME} is evaluated under uniform noise. Feature-dependent noise has not been considered in this setting in literature while it is unlikely that human teachers provide preference with uniform noise.
}

\contribution{
We propose multiple feature-dependent noise models that capture realistic, feature-driven inconsistencies arising from non-expert teachers/LLM feedback.}{
In BPref Benchmark~\citep{lee2021bpref}, only limited types of noise model were benchmarked in PbRL, such as uniform noise, myopic noise, that are not correlated to feature space over trajectory. 
}

\contribution{
We empirically evaluate these noise models using several state-of-the-art PbRL algorithms including a denoising based bechmark to assess their impact on agent learning performance.
}{
None
}


\summary{
Learning from Preferences in Reinforcement Learning (PbRL) has gained attention recently, as it serves as a natural fit for complicated tasks where the reward function is not easily available. However, preferences often come with noise if they are not from perfect teachers. Much prior literature aimed to detect noise, but with limited types of noise and most being uniformly distributed with no connection to observations. In this work, we formalize the notion of targeted feature-dependent noise and propose several variants, such as trajectory feature noise, trajectory similarity noise, margin dependent noise, and language model noise.

We evaluate feature-dependent noise, where noise is correlated with certain feature subsets in complex continuous control tasks from DMControl and Meta-World. Our experiments show that in some feature-dependent noise settings, the state-of-the-art noise-robust PbRL method's learning performance is significantly deteriorated, while a PbRL method with no explicit denoising can surprisingly outperform noise-robust PbRL in the majority of settings. We also find language models' noise exhibits similar characteristics to feature-dependent noise, thereby simulating realistic humans and calling for further study in learning with feature-dependent noise robustly.
}

\begin{document}

\maketitle  %

\begin{abstract}
Learning from Preferences in Reinforcement Learning (PbRL) has gained attention recently, as it serves as a natural fit for complicated tasks where the reward function is not easily available. However, preferences often come with uncertainty and noise if they are not from perfect teachers. Much prior literature aimed to detect noise, but with limited types of noise and most being uniformly distributed with no connection to observations. In this work, we formalize the notion of targeted feature-dependent noise and propose several variants like trajectory feature noise, trajectory similarity noise, margin dependent noise, and Language Model noise. 
We evaluate feature-dependent noise, where noise is correlated with certain features in complex continuous control tasks from DMControl and Meta-world. Our experiments show that in some feature-dependent noise settings, the state-of-the-art noise-robust PbRL method's learning performance is significantly deteriorated, while PbRL method with no explicit denoising can surprisingly outperform noise-robust PbRL in the majority of settings.
We also find language models' noise exhibits similar characteristics to feature-dependent noise, thereby simulating realistic humans and call for further study in learning with feature-dependent noise robustly.
\end{abstract}

\IfFileExists{chapters/introduction.tex}{\input{chapters/introduction}}{
\section{Introduction}
\textbf{Missing file:} \texttt{chapters/introduction.tex}. Upload it (or adjust the path) to include the introduction.
}

\IfFileExists{chapters/preliminary.tex}{\input{chapters/preliminary}}{
\section{Preliminaries}
\textbf{Missing file:} \texttt{chapters/preliminary.tex}. Upload it (or adjust the path) to include this section.
}

\IfFileExists{chapters/methodology.tex}{\input{chapters/methodology}}{
\section{Methodology}
\textbf{Missing file:} \texttt{chapters/methodology.tex}. Upload it (or adjust the path) to include this section.
}

\IfFileExists{chapters/experiments.tex}{\input{chapters/experiments}}{
\section{Experiments}
\textbf{Missing file:} \texttt{chapters/experiments.tex}. Upload it (or adjust the path) to include this section.
}

\IfFileExists{chapters/related_work.tex}{\input{chapters/related_work}}{
\section{Related Work}
\textbf{Missing file:} \texttt{chapters/related_work.tex}. Upload it (or adjust the path) to include this section.
}

\IfFileExists{chapters/conclusion.tex}{\input{chapters/conclusion}}{
\section{Conclusion}
\textbf{Missing file:} \texttt{chapters/conclusion.tex}. Upload it (or adjust the path) to include this section.
}

\bibliography{main}
\bibliographystyle{rlj}

\appendix

\IfFileExists{chapters/appendix.tex}{\input{chapters/appendix}}{
\section{Appendix}
\textbf{Missing file:} \texttt{chapters/appendix.tex}. Upload it (or adjust the path) to include the appendix.
}

\end{document}

%% file: chapters/introduction.tex
\section{Introduction}

Deep Reinforcement Learning (RL) has been successful in recent times and has been deployed extensively in interesting applications covering chip design~\citep{mirhoseini2021graph}, water management systems~\citep{janjua2024gvfs}, gaming companions~\citep{wurman2022outracing} and healthcare~\citep{lakhan2023drlbts}. Despite its success, specifying informative reward functions for RL remains challenging, and they are usually defined by experts or RL developers. There is evidence in literature~\citep{booth2023perils} that reward functions designed by trial and error can often overfit to a specific RL algorithm or learning context and can significantly reduce the overall task metric performance. Proxy reward functions can also lead to unwanted phenomena like reward hacking~\citep{amodei2016concrete,pan2022effects}.

An easier way to specify a reward function is by making it sparse; i.e., provide a reward of $+1$ when a task is completed and $0$ otherwise. Deep RL has been known to suffer from the well-known sample-inefficiency problem due to such sparse reward~\citep{ibarz2021train}, thus making it hard for the agent to learn efficiently. In order to reduce the dependency on hand-crafted reward functions, Preference-based RL (PbRL)~\citep{christiano2017deep,lee2021pebble} has been a popular teacher-in-the-loop paradigm where a reward function is learned from teacher-provided binary preference over pairs of trajectory segments. The Deep RL agent uses the learned reward function to learn an optimal policy well aligned with the teacher's task preference. While generally, these methods have been successful on complex continuous control tasks, they assume access to an oracle for preference labels, which is a limiting assumption.

To address this limiting assumption of access to oracle for preference labels, Lee et al.~\citep{lee2021bpref} introduced various kinds of teachers, including myopic and mistake-scripted teachers, trying to simulate human teachers prone to error.
In this work, we formalise the idea of \textit{feature-dependent noise} within the framework of preference-based RL, motivated by different ways in which humans are prone to error while trying to give comparative feedback on pairs of trajectories. Let us take the example of Figure~\ref{fig:motivation}, where within PbRL, a human teacher encounters two similar trajectories as shown in example E1. It is hard for humans to provide comparative feedback on such trajectories, likely making them prone to error. Another analogous example, as shown in example E2, is when the two sampled trajectories have minute but non-trivial differences (the soccer ball in the figure is barely visible), thus making the teacher skip these important details and hence inducing noise in the preference labels. Prior work ~\citep{lee2021bpref} handles similar trajectory pairs by assigning a neutral preference. However, in practical settings, non-expert annotators may not reliably recognize such similarities, leading to inconsistent or noisy preference feedback.

In this work, we introduce several teacher models of \textit{feature-dependent noise}, which provide practical ways of modelling preference from non-expert teachers. The intuition behind feature-dependent noise is that these noise models depend on specific feature subsets or representations and hence vary as a function of features. These kinds of noise functions arise from uncertainty in human judgment, which is systematically linked to the observable features of trajectories. As an example, if a human teacher induces noise over the preference label because of similarity between the trajectory pair, feature-dependent noise varies as a function of the similarity measure between the two trajectories, which means that the non-expert teacher makes more errors for similar trajectories and less for diverse trajectories. We also empirically evaluate the noise function of language models when they are employed as teachers inside PbRL to understand if they are behaviorally similar to feature-dependent noise.

In recent years, several state-of-the-art algorithms~\citep{Cheng2024RIME, huang2025trend} have proposed denoising mechanisms to identify and filter noisy preference data. In this work, we evaluate feature-dependent noise models using one such state-of-the-art approach. However, because feature-dependent noise is correlated with trajectory features, it is often challenging for these algorithms—designed primarily to handle uniform (feature-independent) noise—to detect such errors effectively. While uniform noise affects preference labels randomly, and is thus more easily identified by existing denoising methods, feature-dependent noise exhibits structured correlations that make it substantially harder to identify and filter, thus leading to poor agent performance.

 \noindent \textbf{Contributions} of this work include (1) formalisation of feature-dependent noise within the PbRL framework, providing a foundation for structured, feature-correlated uncertainty in preference data, (2) introduction of multiple feature-dependent noise models that capture realistic, feature-driven inconsistencies arising from non-expert human/LLM feedback; and (3)  evaluation of these noise models using several state-of-the-art PbRL algorithms to assess their impact on agent learning performance. We introduce and systematically evaluate several feature-dependent noise models for existing PbRL algorithms, which form the main contribution of this work. Evaluation involving VLM-based PbRL is included solely to illustrate the similarity between advice generated by VLMs and feature-dependent noise, and is not the primary focus of this work.
 Extensive experiments on complex continuous control benchmarks from DMControl and Meta-world reveal that these noise functions remain difficult for existing denoising algorithms to detect, thus identifying the need for research in this direction.

%% file: chapters/preliminary.tex
\section{Preliminaries}
\noindent \textbf{Reinforcement Learning:} Reinforcement learning (RL) is represented using a Markov Decision Process (MDP),
which is denoted by $M= ( \mathcal{S}, \mathcal{A}, \mathcal{P}, R, \gamma )$, where $\mathcal{S}$ denotes the agent's state space,  $\mathcal{A}$ is the agent's action space, $\mathcal{P}:\mathcal{S}\times \mathcal{A} \times \mathcal{S} \rightarrow [0,1]$ is the environmental dynamics transition probability, $R: \mathcal{S} \times \mathcal{A} \times \mathcal{S} \rightarrow \mathbb{R}$ is the reward function, and $\gamma$ is a discount factor. The agent's goal is to learn a policy $\pi(a|s)$ which maximizes the discounted sum of rewards.\\
\noindent \textbf{Preference-based RL:} In Preference-based RL (PbRL)~\citep{christiano2017deep}, the agent does not have access to the environment reward function $R$, which is trained from the teacher's preferences. Preferences are binary signals between two trajectory segments, which provide comparative feedback denoting which trajectory segment is favored over another. Given a pair of trajectories, $\tau_1 = \{(s_t^1, a_t^1)\}_{t=0}^T$ and $\tau_2 = \{(s_t^2, a_t^2)\}_{t=0}^T$, 
the preference label  $y \in \{1,0.5,0\}$ denotes whether $\tau_1 \succ \tau_2 (y=1)$~;~ $\tau_1 \prec \tau_2 (y=0)$ ~or~ $\tau_1 = \tau_2 (y=0.5) $. The primary goal in PbRL is to learn a reward model $\hat{R}_\theta(s, a)$, parameterised by $\theta$, that is consistent with preferences. This is done via modelling preferences using the Bradley-Terry model~\citep{bradley1952rank} as below:
\begin{equation*}
    P_{\theta}(\tau_1 \succ \tau_2)=\frac{e^{\sum_t \hat{R}_\theta(s_t^1,a_t^1)}}{e^{\sum_t \hat{R}_\theta(s_t^2,a_t^2)
    }+e^{\sum_t \hat{R}_\theta(s_t^1,a_t^1)}}
\end{equation*}
where $P(\tau_1 \succ \tau_2)$ denotes the probability of preferring trajectory $\tau_1$ over $\tau_2$. Cross-entropy loss between the preference labels and the predicted labels is minimized to update the Reward function $\hat{R}_\theta (s,a)$ as below:
\begin{equation*}
    L(\theta)=-\mathbb{E}[y\log P_\theta (\tau_1 \succ \tau_2)+(1-y)\log P_\theta (\tau_2 \succ \tau_1)]
\end{equation*}

%% file: chapters/methodology.tex
\section{Feature Dependent Noise}\label{sec:feature_dependent_noise}

In this section, we formalize feature-dependent noise (FDN) induced by a sub-optimal teacher in PbRL. We only consider distinct preferences that provide maximum information and filter out ambiguous preferences (y=0.5) as they do not provide an additional discriminative signal during training. We use the same setup as PbRL.

Let $Y$ and $Y^*$ denote random variables for the observed preference label and the unobserved ground-truth label. 
We denote preferences as $y \in \mathcal{Y}$, which represents the teacher's preference over a pair of feature subsets
$\langle \mathbf{X}_1, \mathbf{X}_2 \rangle$, where $\mathbf{X}_1, \mathbf{X}_2 \in \mathcal{P}(\mathbf{X})$, i.e., 
$\mathbf{X}_1$ and $\mathbf{X}_2$ belong to the power set of $\mathbf{X}$ denoted by $\mathcal{P}(\mathbf{X})$.
In PbRL, the feature space $\mathbf{X}$ refers to a feature mapping $\phi: \mathcal{T} \rightarrow \mathcal{P}(\mathbf{X})$
over the states and actions in a trajectory space $\mathcal{T}$.
Given an unobserved ground truth reward function $R_o$, the true trajectory reward over any trajectory $\tau \in \mathcal{T}$ is $G(\tau)=\sum_{i=0}^T \gamma^{i}R_o(s_i,a_i,s_{i+1})$. For each trajectory pair $(\tau_1, \tau_2)$ , we have different types of teachers $T: \langle \mathcal{T}_1,\mathcal{T}_2 \rangle \rightarrow [0,1]$, which is a map from a pair of trajectories to the probability over $\tau_1 \succ \tau_2$.
We define an oracle teacher $T_o$ that gives ground truth preferences $y^*$ as below: 
\begin{equation}
    T_o(\tau_1 \succ \tau_2) = \sigma\left(G(\tau_1) - G(\tau_2)\right)
\end{equation}
 based on the ground truth reward function $R_o$. Here, $\sigma(\cdot)$  is the sigmoid function.
Note that we can also make the oracle teacher deterministic as below:
\begin{equation}
    T_o(\tau_1 \succ \tau_2) = 
\begin{cases}
1 & \text{if } G(\tau_1) > G(\tau_2) , \\
0  & \text{if } G(\tau_1) < G(\tau_2).
\end{cases}
\label{equ:deterministic_teacher}
\end{equation}

To model a non-expert teacher $T_n$, we define a noise function 
$N(\tau_1, \tau_2): \mathcal{T}^2 \rightarrow [0,1]$, representing the probability 
of mistakenly flipping a preference label given a trajectory pair 
$(\tau_1, \tau_2)$ and unobserved ground truth preference label $y^*$.Mathematically, the noise function 
$N(\tau_1,\tau_2) = P(y \neq y^* \mid y^*, \phi(\tau_1), \phi(\tau_2))$ 
is defined over feature subsets corresponding to the trajectory pairs. The model of the  non-expert teacher is represented as:
\begin{equation}
    T_n(\tau_1 \succ \tau_2) = \underbrace{T_o(\tau_1 \succ \tau_2)(1-N(\tau_1, \tau_2))}_{\text{correct\ preference score}} + \underbrace{T_o(\tau_2 \succ \tau_1)N(\tau_1, \tau_2)}_{\text{noisy\ preference score}} 
\end{equation}
In the above equation, the first part represents the probability that the noisy teacher $T_n$ chooses the preference label correctly in accordance with the ground truth (oracle teacher), and the second part denotes the probability that it chooses the trajectory ordering $(\tau_1 \succ \tau_2)$ incorrectly, opposite to the ground truth label.
The assumption is that the noise function is symmetric so $\forall \tau_1, \tau_2, N(\tau_1, \tau_2)=N(\tau_2, \tau_1)$.
While this function can be a constant, i.e., $N(\tau_1, \tau_2)=C$, modelling uniform distribution noise, we focus on more complicated cases, where this probability depends on the feature space, giving us feature-dependent noise.

\begin{figure*}
    \centering
    \includegraphics[width=1\linewidth]{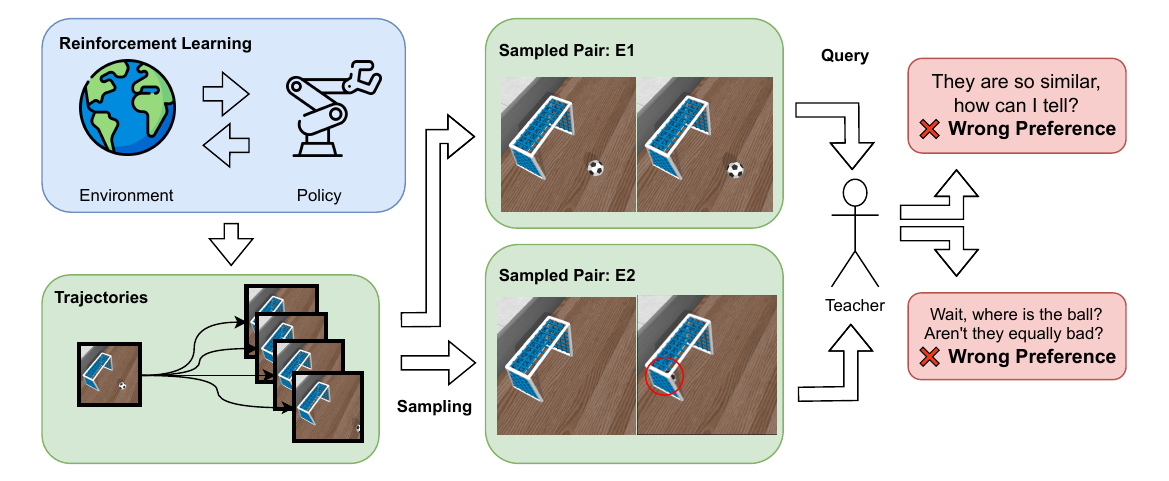}
    \caption{\small Examples of feature-dependent noise. A teacher may be prone to errors because of similarities (E1) or hidden details in the observation that are hard to notice (E2). We explore more types of FDN in our experiments.}
    \label{fig:motivation}
\end{figure*}

\subsection{Feature Dependent Noise Categories}\label{sec:categories}
In this section, we discuss various types of Feature-Dependent Noise that follow from the FDN formalization in the previous section.

\noindent \textbf{Trajectory Similarity Noise} The intuition behind this noise is that if two trajectories are similar, the probability of inducing FDN by teachers increases and vice-versa.
In Trajectory Similarity Noise, the feature is a pair of full trajectories $x=(\phi(\tau_1),\phi(\tau_2))$  and the noise function is $N(\tau_1,\tau_2) \sim \frac{1}{D(\phi(\tau_1),\phi(\tau_2))}$, where $D$ is a distance measure. In our settings, we consider the whole trajectory so that $\phi$ is an identity mapping.
An instance could be, $N(\tau_1,\tau_2) = min(1, \frac{1}{||\phi(\tau_1)-\phi(\tau_2))||^2_2})$, where the probability of noise is proportional to the L2 distance between two trajectories. 
Another way of computing D would be to use 
encoders to compute distance in latent trajectory space, where $D =||\phi(E(\tau_1))-\phi(E(\tau_2))||^2_2$; $E(\tau)$ refers to the encoder function that outputs trajectory representation in latent space.\\

\noindent \textbf{Trajectory Feature Magnitude Noise:} Human teachers often struggle to reliably distinguish between trajectories when the differences are concentrated in certain feature subsets that strongly affect perceived stability. In particular, in robotics domains, large variations in the torques applied across joints can cause the resulting trajectories to appear visually unstable. This instability increases the likelihood of inducing FDN.
In this type of noise, the feature is a subset of the trajectory features. These features are predefined from domain knowledge, where the teacher lacks the ability to distinguish good or bad trajectory segments owing to high variations (change in magnitude) of the feature subsets.
The feature magnitude noise is defined over a pair of trajectories features $x = (\phi(\tau_1),\phi(\tau_2))$, where each
trajectory is summarized by the time-averaged norm of its 
state or action feature subsets. Here, the feature mapping $\phi$ maps to a subset in the feature space $\mathbf{X}$. Let $\Delta = \lVert \phi(\tau_1) \rVert - \lVert \phi(\tau_2) \rVert$, where $\lVert \phi(\tau) \rVert = \tfrac{1}{T}\sum_{t=1}^{T}\lVert \phi(\tau)_t \rVert_2$ denotes the mean L2-norm over a feature subset as per the trajectory. The noise function is defined as $N(\tau_1, \tau_2) = \sigma \left( \beta \, \log\!\big(1 + |\Delta|\big) \, \mathrm{sign}(\Delta) \right)$
where $\beta$ is a scaling parameter.
The scaling parameter controls how aggressively the noise reacts to the feature magnitude difference. Small $\beta$ denotes weak FDN and vice versa. We used $\beta$=1 for the experiments. The sign function $\Delta$ denotes the difference in  magnitudes of trajectory feature subsets, such that $N(\tau_1,\tau_2)$
increases when one trajectory exhibits larger feature magnitudes compared to the other and vice versa. A Bernoulli sample from $N(\tau_1,\tau_2)$ determines whether the observed preference label is flipped or not, thus inducing FDN.\\

\noindent \textbf{Margin-dependent Noise:}
We construct an FDN model in which the probability of label noise increases as the distinguishability between trajectory returns decreases. Under the Bradley–Terry model, the preference probability over a feature subset of trajectory pair $x = (\phi(\tau_1), \phi(\tau_2))$ depends on the difference in predicted returns as below:
\[
P_\theta(\phi(\tau_1)\succ\phi(\tau_2) )
=
\sigma\!\left(\Delta \hat{G}_\theta(\phi(\tau_1),\phi(\tau_2)\right),
\]
where 
\[
\Delta \hat{G}_\theta(\phi(\tau_1),\phi(\tau_2))
=
\sum_{t} \hat{R}_\theta(s_t^{(1)}, a_t^{(1)})
-
\sum_{t} \hat{R}_\theta(s_t^{(2)}, a_t^{(2)}).
\]

When $|\Delta \hat{G}_\theta(x)|$ is small, the induced preference probability approaches $0.5$, corresponding to maximal ambiguity between the trajectories.
We therefore define a margin  score $m(x) = \left| \Delta \hat{G}_\theta(\phi(\tau_1),\phi(\tau_2)) \right|^{-1}$,
or equivalently, any monotonic transformation of the model confidence.
Trajectory pairs are ranked according to $m(x)$, and the top $\epsilon\%$ most uncertain samples are selected for label flipping, where $\epsilon$ controls the overall corruption rate.
The noise function $N(\tau_1,\tau_2)$ is defined as:
\begin{equation}
N(\tau_1, \tau_2) =
\begin{cases}
1 & \text{if } |\Delta \hat{G}_\theta(\phi(\tau_1),\phi(\tau_2))| < t, \\
0 & \text{otherwise},
\end{cases}
\label{equ:uncertainty}
\end{equation}
where the threshold $t$ is chosen such that an $\epsilon$ fraction of samples satisfy the condition.
The induced FDN model is model-relative and can be interpreted as hypothesis-dependent feature noise. In particular, corruption is applied to samples satisfying $|\Delta \hat{G}_\theta(\phi(\tau_1),\phi(\tau_2))| < t$, corresponding to points near the decision boundary of the learnt reward model. As the absolute margin $|\Delta \hat{G}_\theta(\phi(\tau_1),\phi(\tau_2))|$ increases, samples lie farther from the boundary and are therefore less subject to corruption. This construction yields structured feature-dependent noise that adapts to the hypothesis class and concentrates corruption in regions where the model is least confident (most uncertain).

\noindent \textbf{Hybrid Noise:}
The intuition behind hybrid noise is that some samples may be ambiguous due to both behaviorally small differences (similar trajectories) or instability (high feature subset magnitude) and low model confidence (as indicated by similar returns). Hence, in this type of FDN, we combine the noise model of behavioral FDN with margin dependent noise. Targeted behavioral noise in areas where the reward model is highly ambiguous, which would result in an induced noise distribution correlated with the true preference distribution and hence make it difficult for the reward model to distinguish between preference ambiguity and preference annotation error. For each trajectory
pair $x = (\tau_1,\tau_2)$, we compute a trajectory behavior-based  score denoted by
$\text{score}_{\text{f}}(x)$ and a margin-uncertainty-based score denoted by
$\text{score}_{\text{u}}(x)$ derived from marginal uncertainty of the
reward model. For $\text{score}_{\text{f}}(x)$, it can be any kind of other noise. For example, we can take trajectory distance as $\text{score}_{\text{f}}(x)$, giving a hybrid noise of trajectory similarity noise and margin dependent noise.
The total score  for every trajectory pair is
\begin{equation*}
\text{score}(x) = \alpha \cdot \text{score}_{\text{f}}(x) + (1-\alpha)\cdot \text{score}_{\text{u}}(x),
\end{equation*}
Here, $\alpha \in [0,1]$ is a weight coefficient that balances the contribution of the feature-based score and the margin-uncertainty score. Here, if $\alpha=0$, then it gives margin dependent noise and if $\alpha=1$, it gives the feature-based noise. \\

\noindent \textbf{Language Model Noise:} In this type of noise, we employ an LLM or VLM as a teacher for eliciting preference, as done in prior work like RL-VLM-F~\citep{wang2024} and RL-SaLLM-F~\citep{tu2024online}. LLMs are inherently known for providing noisy advice by virtue of properties like hallucination~\citep{xu2024hallucination}. The judgments of language models majorly rely on latent representations rather than true reward signals. They are often biased towards salient or easily perceived feature subsets rather than task-relevant dynamics; hence, the induced noise is most likely to be an FDN. The goal here is to employ a language model as a teacher to deduce if the noisy distribution induced by these models is closer to FDN and hence difficult to detect by existing denoising techniques in PbRL literature.

%% file: chapters/experiments.tex
\setkeys{Gin}{keepaspectratio}
\newcommand{\envimg}[1]{\includegraphics[height=3.5cm]{figures/domains/#1}}
\section{Experiments}
    The experiments are designed to answer the following research questions:
    \begin{itemize}[label={}, leftmargin=*]
    \item \textbf{R1.}  Can current state-of-the-art PbRL denoising method, RIME effectively handle feature-dependent noise? 
    \item \textbf{R2.} How do the proposed variants of feature-dependent noise compare against each other within the PbRL framework?
    \item\textbf{R3.} Do LLMs induce feature-dependent noise?
    \item\textbf{R4.} 
    How do state-of-the-art PbRL algorithms without a denoising component compare to RIME in handling FDN?

    \end{itemize}
\subsection{Experiment Setup}
We follow the general experimental design from RIME~\citep{Cheng2024RIME}, adapting it to study feature-dependent noise rather than only uniform noise. Specifically, we evaluate on three locomotion domains from DMControl~\citep{tassa2018dmcontrol}
: \textbf{Walker}, \textbf{HalfCheetah}, and \textbf{Quadruped}. These tasks provide diverse control dynamics and allow us to test noise sensitivity across environments. We also have experiments from Meta-World on Hammer, Sweep-Into and Button Press as reported in the Appendix Section~\ref{app:meta_world}.
A scripted teacher provides pairwise trajectory preferences based on ground-truth episodic returns as Equation~\ref{equ:deterministic_teacher}, which are then corrupted according to the noise models defined in Section~\ref{sec:categories}. We inject noise rates of \textbf{10\%, 20\%, 30\%, and 40\%}, consistent with robustness studies in prior work. 

For a fair comparison, we follow RIME’s preference-based RL setup. 
Walker and HalfCheetah 
uses $1000$ preference queries at every learning step and a reward batch size of $100$. 
Quadruped, being more challenging, uses
$4000$ preference queries per learning step and a reward batch of $400$. 
For all the environments, we use unsupervised pre-training to pre-train the reward model as done in RIME.
All results are averaged over 5 runs, and the mean episodic return and standard deviation are reported.\footnote{If the denoising algorithm reports poor agent performance with feature-dependent noise rather than uniform noise, then it is harder to detect.} We use \textbf{RIME}~\citep{Cheng2024RIME} as a baseline, as it is the current state-of-the-art method in PbRL to detect and filter uniform random noise over preference labels. More details are shown in Appendix Section~\ref{app:exp_settings}. We also evaluated other PbRL algorithms such as SURF~\citep{DBLP:conf/iclr/ParkSSLAL22}, RUNE~\citep{liang2022reward}, PEBBLE~\citep{lee2021pebble} in Appendix. We did not benchmark TREND~\citep{huang2025trend} as no open-source code is available for replication.

\begin{figure*}[t]
    \includegraphics[width=1.00\linewidth]{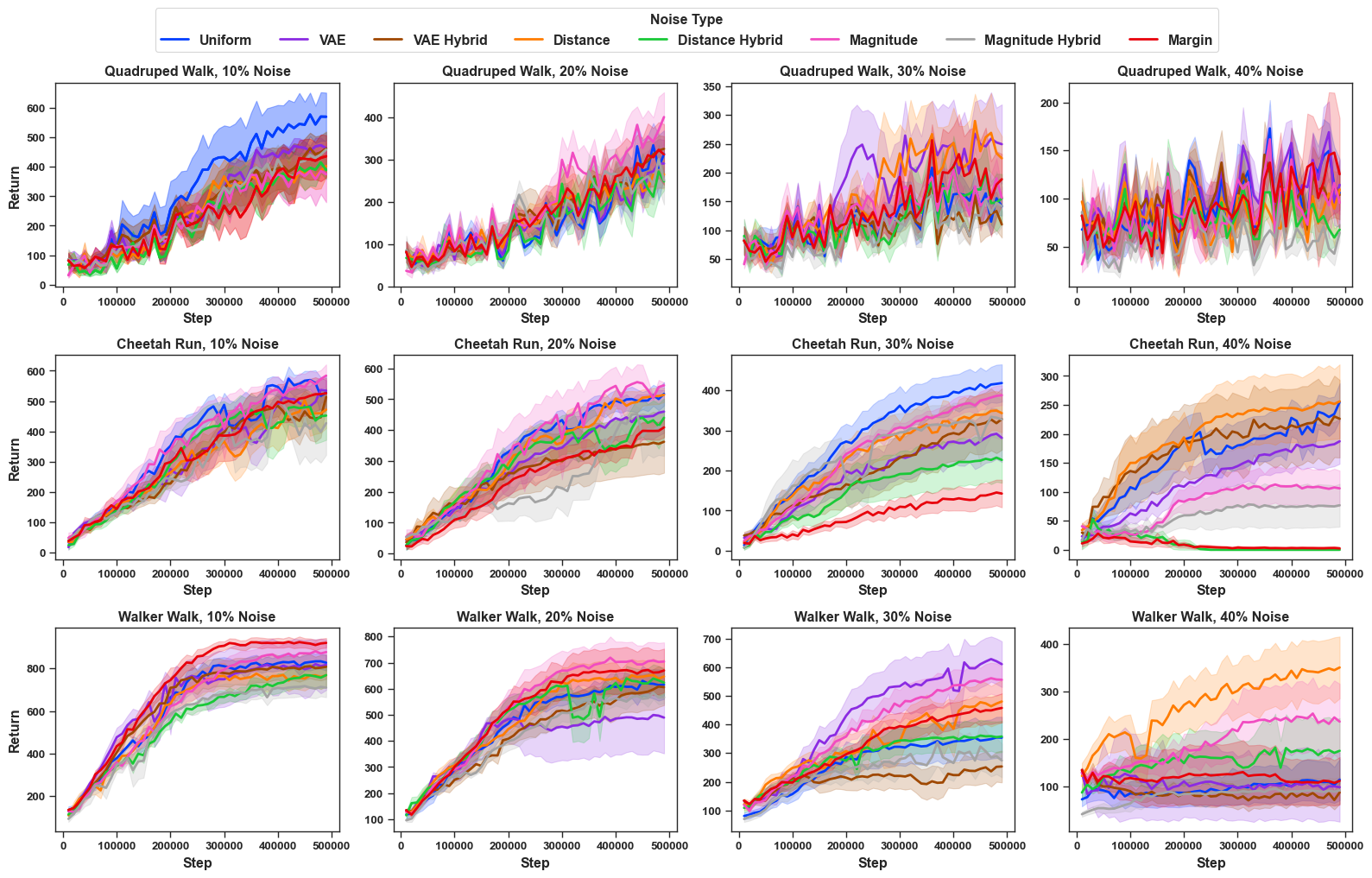}
  \caption{ \small
    Each row is a domain:
    Walker-walk, HalfCheetah-run, Quadruped-walk. 
    Curves show mean $\pm$ standard error over seeds; x-axis is \textit{Step},
    y-axis is \textit{Episodic return}.} 
  \label{fig:comapre_noise}
\end{figure*}

\subsection{Results and analysis}

\noindent \textbf{Trajectory Feature Magnitude Noise:}
This noise flips labels for trajectory pairs that show big differences in their action-level features. In this setup, we use the \textit{torque magnitudes from the action space} of each domain to define the noise. The results are presented in Figure~\ref{fig:comapre_noise} as denoted by \textbf{Magnitude} (pink).

In Walker, Trajectory Feature Magnitude Noise is slightly easier to detect than Uniform noise at 10\% with better agent performance than Uniform (blue line) and becomes increasingly easier to detect than Uniform at higher noise rates. In Half Cheetah, this noise is similar in performance to Uniform noise at 10-20\% corruption but is harder to detect than Uniform noise at 30-40\%. In Quadruped, this noise is harder to detect at 10\% and easier to detect or comparable at higher noise levels.
The effect of trajectory feature noise is domain- and noise-level-dependent and does not have a clear pattern. We hypothesize that the structure of this type of noise makes it easier to detect; RIME can easily recognize that all samples that have large torque values are noise and hence, are easier to detect than identifying a random noisy sample. This detection becomes even easier at higher noise levels due to the availability of more noisy samples.

\noindent \textbf{Margin Dependent Noise:}
We simulate margin dependent Noise by injecting noise into trajectories where the reward function has the highest marginal uncertainty, i.e., the predicted reward between two trajectories is very close to each other. This noise refers to value-based similar trajectories as seen from the lens of the reward model.
The results are presented in Figure~\ref{fig:comapre_noise} as denoted by \textbf{Margin} (red). It can be seen that this noise is generally harder with lower learning performance in our domains as compared to Uniform noise denoted by the orange line, except for Walker, and the agents tend to converge to a much lower episodic return, sometimes not learning at all (e.g, 30\% noise on Half Cheetah). This suggests that the previous denoise algorithms are still challenged by this type of noise, which is non-stationary and adapts to the evolving learned hypothesis, thereby making it particularly difficult to detect.

\noindent \textbf{Trajectory Similarity Noise}
We tested the trajectory similarity noise with two distance metrics: (1) L2 distance and (2) VAE embedding distance. 
In L2 distance, we take the L2 norm distance between the trajectory pairs. 
In the VAE embedding distance, we pretrain a VAE encoder to embed the trajectory into a much smaller vector representation, and then we take the L2 distance between the two embeddings. We use MLP and transformers as our encoders.
The details of our encoder training and architecture can be found in Appendix Section~\ref{app:vae}. 
The learning curves under trajectory similarity noise can be found in Figure~\ref{fig:comapre_noise} as denoted by \textbf{Distance} and \textbf{VAE}. 
It is observed that the VAE (purple line) can be significantly harder in comparison with Uniform Noise across domains. For example, VAE noise deteriorates the episodic return in all of our domains under most noise percentages, with Walker under 30\% being an exception.
L2 Distance, on the other hand, shows a trend to be easier to handle, and the policy still learns relatively well against up to 40\% noise in Walker and Cheetah. One reason for this is that similar trajectories come with similar rewards, and wrong preferences over similar trajectories usually provide a smaller penalty to reward function learning, while this might not hold for latent space representation.

\noindent \textbf {Hybrid Noise:}
Here, we combine two criteria: how uncertain the reward model is about a preference, and how similar or unstable the trajectories are under chosen features. 
The weight coefficient $\alpha$ is a hyperparameter to determine the contribution of individual noise functions.
We study two types of hybrid noise:
\begin{itemize}
  \item[1.] \emph{Magnitude Hybrid Noise}: targets pairs with large contrast in feature magnitudes when the model is also uncertain.
  \item[2.] \emph{Similarity Hybrid Noise}: targets pairs that look behaviorally alike (e.g., small distances in feature or embedding space) with high model uncertainty.
\end{itemize}

\noindent \textbf{Magnitude Hybrid Noise:} 
This noise reflects a challenging pattern, as the teacher provides incorrect feedback on samples where the reward function is uncertain, and the teacher is erroneous due to behavioral instability. The teacher makes mistakes in feature space, where the reward model is most likely to get preferences wrong. Results are shown in Figure~\ref{fig:comapre_noise} denoted by \textbf{Magnitude Hybrid} (grey).
In HalfCheetah (Figure~\ref{fig:comapre_noise}a–d), Magnitude Hybrid Noise performs almost the same as  Uniform at 10–20\% corruption ($\alpha=0.9$ at 10\%, $\alpha=0.3$ at 20\%) but shows stronger performance at higher noise levels. At 30\% ($\alpha=0.3$), the results demonstrate that Magnitude Hybrid performs better than Uniform noise. At 40\% ($\alpha=0.1$), the results show that the algorithm fails to learn because aggressive flipping in ambiguous regions causes collapse, while Uniform still retains some learning ability. 
In Walker, Magnitude Hybrid demonstrates slightly better performance than Uniform at 10–20\% ($\alpha=0.5$ at 10\% and 20\%). At 30-40\% ($\alpha=0.7$ at 30\%, $\alpha=0.9$ at 40\%), Magnitude Hybrid noise performs significantly better, severely degrading agent performance by targeting highly uncertain preference pairs. 
In Quadruped, the advantage of Magnitude Hybrid (harder to detect and hence, lower performance) emerges clearly at all noise levels ($\alpha=0.9$ at 10\%, $\alpha=0.5$ at 20\% and 30\%, $\alpha=0.3$ at 40\%). Unlike Walker, the difficulty of this noise remains consistent.\\
To summarize, Magnitude Hybrid noise on average, is harder to detect than pure Trajectory Feature Noise consistently in every domain.

\noindent \textbf{Similarity Hybrid Noise:}
We also test Hybrid Noise from Margin Dependent Noise and trajectory similarity noise (both L2 and VAE). We take $\alpha=0.5$ for these experiments. As shown in Figure~\ref{fig:comapre_noise} denoted by \textbf{Distance Hybrid}(green) and \textbf{VAE Hybrid}(brown), this fusion makes noise much more challenging to learn from compared to Uniform noise. With the increase in noise ratio, all types of noise become hard to tackle, and this effect is most significant in low-scale noises, as a high proportion of noise, regardless of the type of noise, generally flattens the learning curve. For example, under 10\% noise, hybrid noise gives a lower episodic return in all three domains in comparison with Uniform noise. We also observe that the Distance Hybrid learning curve almost flattens under 40\% noise in HalfCheetah, while the Distance noise itself in the same scale still allows satisfactory episodic reward, thus emphasizing the importance of behavioral noise in uncertain areas.

To summarise, we found several hybrid noises that pose a harder challenge to preference-based RL algorithms, and this effect is often more significant under low-scale noise of 10\%, where the proposed FDN is harder to detect (lower agent performance) than uniform noise $83\%$ of the time across all domains in DMControl.\footnote{Refer to  Table~\ref{tab:harder_than_uniform_dmcontrol} and  ~\ref{tab:harder_than_uniform_metaworld} in appendix for more summary statistics of FDNs on DMControl and Metaworld domains.} Here to answer \textbf{R1} and \textbf{R2}, the current state-of-the-art PbRL denoising methods cannot handle FDN effectively as compared to uniform noise. In comparison with trajectory similarity noise or trajectory feature magnitude noise, hybrid noise often renders as the most challenging one to filter by RIME. Table~\ref{tab: episodic_return_all_noise} reports the final mean return for all noise levels across all domains, demonstrating that some variant of hybrid noise outperforms other variants approximately $70\%$ of the time, thereby supporting the claim.

\begin{table*}[t]
\centering
\label{tab:3d_results}

\renewcommand{\arraystretch}{0.9}
\setlength{\tabcolsep}{3pt}
\scriptsize

\resizebox{0.78\textwidth}{!}{
\begin{tabular}{lcccc}
\toprule
\textbf{Noise Type} & \textbf{10\%} & \textbf{20\%} & \textbf{30\%} & \textbf{40\%} \\
\midrule

\multicolumn{5}{l}{\textbf{Walker Walk}} \\[1pt]
\hspace{6pt} Uniform & 847.75 $\pm$ 99.16 & 633.06 $\pm$ 153.14 & 362.65 $\pm$ 177.31 & 119.69 $\pm$ 104.43 \\
\cmidrule(lr){1-5}
\hspace{6pt} Adversarial & 898.05 $\pm$ 83.51 & 821.54 $\pm$ 115.46 & 549.61 $\pm$ 156.66 & 216.74 $\pm$ 112.42 \\
\hspace{6pt} Distance & 776.65 $\pm$ 138.90 & 657.81 $\pm$ 146.75 & 516.84 $\pm$ 227.59 & 359.93 $\pm$ 166.41 \\
\hspace{6pt} Distance Hybrid & 762.23 $\pm$ 133.60 & 659.51 $\pm$ 139.02 & 368.64 $\pm$ 169.42 & 170.43 $\pm$ 169.81 \\
\hspace{6pt} Magnitude & 905.48 $\pm$ 98.08 & 722.86 $\pm$ 181.32 & 624.71 $\pm$ 94.10 & 271.29 $\pm$ 220.74 \\
\hspace{6pt} Magnitude Hybrid & \textbf{728.32 $\pm$ 97.73} & \textbf{560.08 $\pm$ 33.01} & 337.64 $\pm$ 201.10 & 90.07 $\pm$ 63.68 \\
\hspace{6pt} Margin & 917.94 $\pm$ 64.60 & 673.86 $\pm$ 213.63 & 498.39 $\pm$ 130.01 & 113.40 $\pm$ 124.70 \\
\hspace{6pt} VAE & 883.65 $\pm$ 162.76 & 813.90 $\pm$ 120.60 & 662.56 $\pm$ 172.63 & 94.4 $\pm$ 90.38 \\ 
\hspace{6pt} VAE Hybrid & 828.77 $\pm$ 137.24 & 639.40 $\pm$ 197.79 & \textbf{266.72 $\pm$ 144.05} & \textbf{84.04 $\pm$ 34.37} \\

\midrule
\multicolumn{5}{l}{\textbf{HalfCheetah Run}} \\[1pt]
\hspace{6pt} Uniform & 651.99 $\pm$ 83.67 & 555.41 $\pm$ 78.50 & 473.45 $\pm$ 82.86 & 308.52 $\pm$ 102.11 \\
\cmidrule(lr){1-5}
\hspace{6pt} Adversarial & 564.78 $\pm$ 150.79 & 531.39 $\pm$ 72.42 & 407.81 $\pm$ 138.83 & 25.60 $\pm$ 35.68 \\
\hspace{6pt} Distance & 585.11 $\pm$ 80.76 & 567.91 $\pm$ 64.00 & 380.60 $\pm$ 106.07 & 270.26 $\pm$ 161.26 \\
\hspace{6pt} Distance Hybrid & 562.29 $\pm$ 88.68 & 507.53 $\pm$ 124.04 & 295.37 $\pm$ 131.99 & \textbf{0.04 $\pm$ 0.08} \\
\hspace{6pt} Magnitude & 641.62 $\pm$ 65.04 & 593.65 $\pm$ 72.30 & 440.46 $\pm$ 85.82 & 114.39 $\pm$ 68.17 \\
\hspace{6pt} Magnitude Hybrid & \textbf{522.45 $\pm$ 197.21} & 496.54 $\pm$ 86.36 & 402.65 $\pm$ 50.12 & 79.08 $\pm$ 73.02 \\
\hspace{6pt} Margin & 549.61 $\pm$ 222.15 & 455.43 $\pm$ 136.03 & \textbf{181.93 $\pm$ 113.91} & 2.22 $\pm$ 5.22 \\
\hspace{6pt} VAE & 601.47 $\pm$ 65.76 & 531.71 $\pm$ 176.34 & 368.54 $\pm$ 163.44 & 212.00 $\pm$ 106.56 \\
\hspace{6pt} VAE Hybrid & 569.34 $\pm$ 67.63 & \textbf{409.75 $\pm$ 227.47} & 432.64 $\pm$ 180.67 & 250.95 $\pm$ 150.20 \\

\midrule
\multicolumn{5}{l}{\textbf{Quadruped Walk}} \\[1pt]
\hspace{6pt} Uniform & 575.12 $\pm$ 270.34 & 327.58 $\pm$ 168.10 & 312.93 $\pm$ 194.09 & 102.31 $\pm$ 24.94 \\
\cmidrule(lr){1-5}
\hspace{6pt} Adversarial & 508.28 $\pm$ 227.63 & 431.60 $\pm$ 186.71 & \textbf{84.78 $\pm$ 33.71} & 79.15 $\pm$ 41.79 \\
\hspace{6pt} Distance & 443.33 $\pm$ 134.45 & 278.23 $\pm$ 132.74 & 214.39 $\pm$ 134.53 & 89.31 $\pm$ 52.76 \\
\hspace{6pt} Distance Hybrid & \textbf{326.10 $\pm$ 178.83} & \textbf{231.31 $\pm$ 93.84} & 155.98 $\pm$ 129.45 & 90.70 $\pm$ 46.13 \\
\hspace{6pt} Magnitude & 567.89 $\pm$ 168.26 & 419.24 $\pm$ 123.36 & 216.15 $\pm$ 69.60 & 107.04 $\pm$ 56.45 \\
\hspace{6pt} Magnitude Hybrid & 478.47 $\pm$ 160.87 & 311.65 $\pm$ 125.62 & 129.74 $\pm$ 66.84 & \textbf{58.16 $\pm$ 33.26} \\
\hspace{6pt} Margin & 459.71 $\pm$ 240.22 & 340.46 $\pm$ 180.73 & 194.89 $\pm$ 118.45 & 120.15 $\pm$ 93.05 \\
\hspace{6pt} VAE & 402.37 $\pm$ 274.31 & 316.91 $\pm$ 187.13 & 229.84 $\pm$ 152.31 & 130.21 $\pm$ 45.45 \\
\hspace{6pt} VAE Hybrid & 523.06 $\pm$ 163.73 & 371.02 $\pm$ 98.01 & 150.24 $\pm$ 57.07 & 120.14 $\pm$ 55.54 \\
\bottomrule
\end{tabular}}
\caption{\small Final episodic return (mean $\pm$ std) across domains and noise levels for each noise type. \emph{Uniform} serves as the reference.}
\label{tab: episodic_return_all_noise}
\end{table*}

\noindent \textbf{Language Model Noise:}
We tested with Qwen 2.5 VL series model, with model sizes of 7B, 32B and 72B, as teachers to provide preferences. We tested two visual domains, Cart Pole and Metaworld Soccer. We chose these two domains as they provide intuitive visual signals for preference feedback. 
In CartPole, the goal is to keep the rod vertical to the ground as much as possible, and therefore, the teacher may simply compare the angles of the rod to provide high-quality references.
In Metaworld Soccer, the agent needs to control a robot arm to move the soccer into the gate, and the teacher can provide high-quality preferences by observing the distance between the soccer and the gate. The prompts we use to elicit preference follow similar settings in ~\cite{wang2024} and can be seen in Appendix ~\ref{app:prompt}.

\begin{figure}[t]
  \centering
  \begin{subfigure}[t]{0.48\textwidth}
    \centering
    \includegraphics[width=\linewidth]{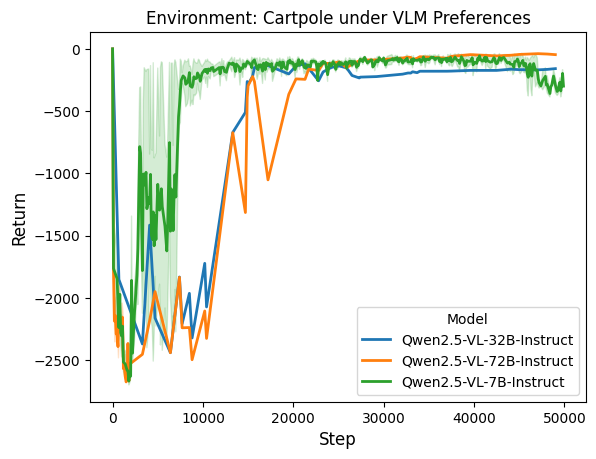}
    \subcaption{CartPole under VLM preferences}
    \label{fig_vlm_cartpole}
  \end{subfigure}
  \hfill
  \begin{subfigure}[t]{0.48\textwidth}
    \centering
    \includegraphics[width=\linewidth]{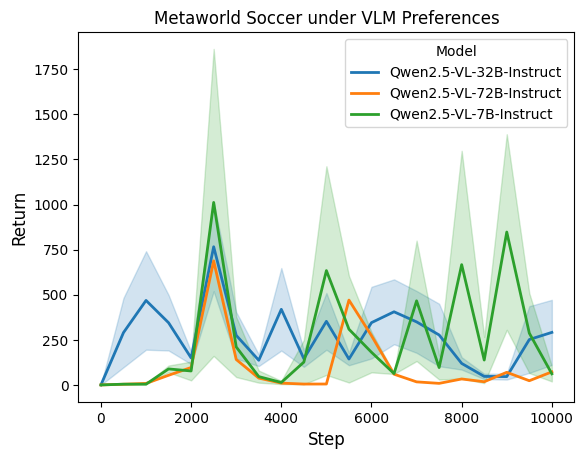}
    \subcaption{Metaworld Soccer under VLM preferences}
    \label{fig_vlm_soccer}
  \end{subfigure}
  \caption{\small Learning performance on VLM-sourced preferences on CartPole and Metaworld Soccer.}
  \label{fig:vlm}
\end{figure}

\begin{table}
    \centering
    \resizebox{0.8\columnwidth}{!}{%

    \begin{tabular}{lcccc}
           &\textbf{Qwen2.5VL-7B} & \textbf{Qwen2.5VL-32B} & \textbf{Qwen2.5VL-72B} \\
         \midrule
         \textbf{CartPole} &  &  &  \\
         \hspace{1em}Noise & 0.458 & 0.070 & 0.008  \\
         \hspace{1em}Return (VLM) & -201.42&-158.31 & -45.96\\
         \hspace{1em}Return (Uniform) & -2237.46&-21.70 & -23.07\\
         \midrule
        \textbf{Metaworld Soccer} &  &  &  \\
        \hspace{1em}Noise & 0.463 & 0.356 & 0.296  \\
        \hspace{1em}Return (VLM) & 5.20& 301.18 & 84.75\\
         \hspace{1em}Return (Uniform) & 66.73&361.54 & 3.33\\
         \bottomrule
    \end{tabular}
    }
    \caption{\small VLM preference noise and episodic returns from different models. We also present episodic returns from the corresponding uniform noise for comparison.}
    \label{tab:vlm_noise}
\end{table}
The results can be seen in Figure~\ref{fig:vlm} and the corresponding noise can be seen in Table~\ref{tab:vlm_noise}\footnote{Due to limited computation, we only show runs with one seed for bigger models.}.
We can find that in the Cart Pole, even the smallest model can achieve a high episodic return. Though with a rather small model like Qwen 2.5 VL 7B, the preference noise reaches as high as $0.458$, the agent is still able to learn against such a high level of noise, while in the same proportion of Uniform noise, we see the learned policy completely failed in the task. This is due to the fact that most errors in preferences are made in similar images. Examples of wrong preferences are presented in Appendix~\ref{app:prompt}. As a result, the learned reward functions are still able to correctly penalise or encourage desired behaviour in most of the observations in Cart Pole. Another observation here is that the smaller VLM (Qwen2.5-7B) drives faster learning than stronger models, indicating that smaller, noisier teachers can still offer more effective feedback—aligning with the larger model's paradox~\cite{xu2025stronger} in literature.

We observed a similar pattern in Metaworld Soccer, where preference errors are often made in similar image observations. However, the Metaworld Soccer is a much more complicated domain that requires 3D understanding, and all of the models fail to provide high-quality preferences. For example, even the biggest model, Qwen 2.5 VL 72B, gives about 29.6\% noise, and our smallest model, Qwen 2.5 VL 7B, almost gives random preferences. As a result, none of the models can guide the policy to solve the task, and at the same scales of uniform noise, similarly, all policies failed to complete the task. 
Furthermore, sometimes the soccer ball is almost blocked by the gate, and the VLM may not notice it, just like a human teacher. 
Here, to answer our research question \textbf{R3}, VLM-induced noise exhibits characteristics similar to trajectory-similarity noise, suggesting that VLM-based teachers inherently introduce feature-dependent noise.
Learning with this type of noise is challenging in complex high-dimensional domains like Meta-world Soccer as opposed to an easy domain like Cartpole.

\noindent \textbf{Different PbRL algorithms under FDN:}
To answer R4, we further benchmarked the learning performance of other PbRL algorithms that do not explicitly handle noisy preference, including PEBBLE~\citep{lee2021pebble}, SURF~\citep{DBLP:conf/iclr/ParkSSLAL22} and RUNE~\citep{liang2022reward}. We compare them under a fixed setting—Cheetah Run with different scales and all eight types of noise—as shown in Figure~\ref{fig:compare_algo_0.2}, with other results in Appendix Section~\ref{app:other_algo} as in Figure~\ref{fig:compare_algo_0.1}, Figure~\ref{fig:compare_algo_0.3}, and Figure~\ref{fig:compare_algo_0.4}. Overall, SURF (denoted in green) often performs the worst among the four methods. A plausible explanation is that SURF augments preference labels using its learned reward model, which could amplify label errors when the teacher is imperfect. In contrast, RUNE tends to be the most stable. RIME shows substantial variability across noise types; for instance, under high distance-hybrid noise (30\% and 40\%), it can even perform worse as compared to other algorithms. We also observe in the majority cases (94\% cases in Cheetah run; 63\% in Walker walk; 56\% in quadruped),  RIME is not consistently the best in terms of performance. This result further highlights the inherent difficulty of feature-dependent noise, where an explicit denoising method may fail to generalize and can underperform as compared to other non-denoising methods.
Additional results of individual comparison of PEBBLE, SURF and RUNE under different noise models can be found in Appendix ( Section~\ref{app:other_algo}).

\begin{figure*}[h]
    \centering
    \includegraphics[width=1\linewidth]{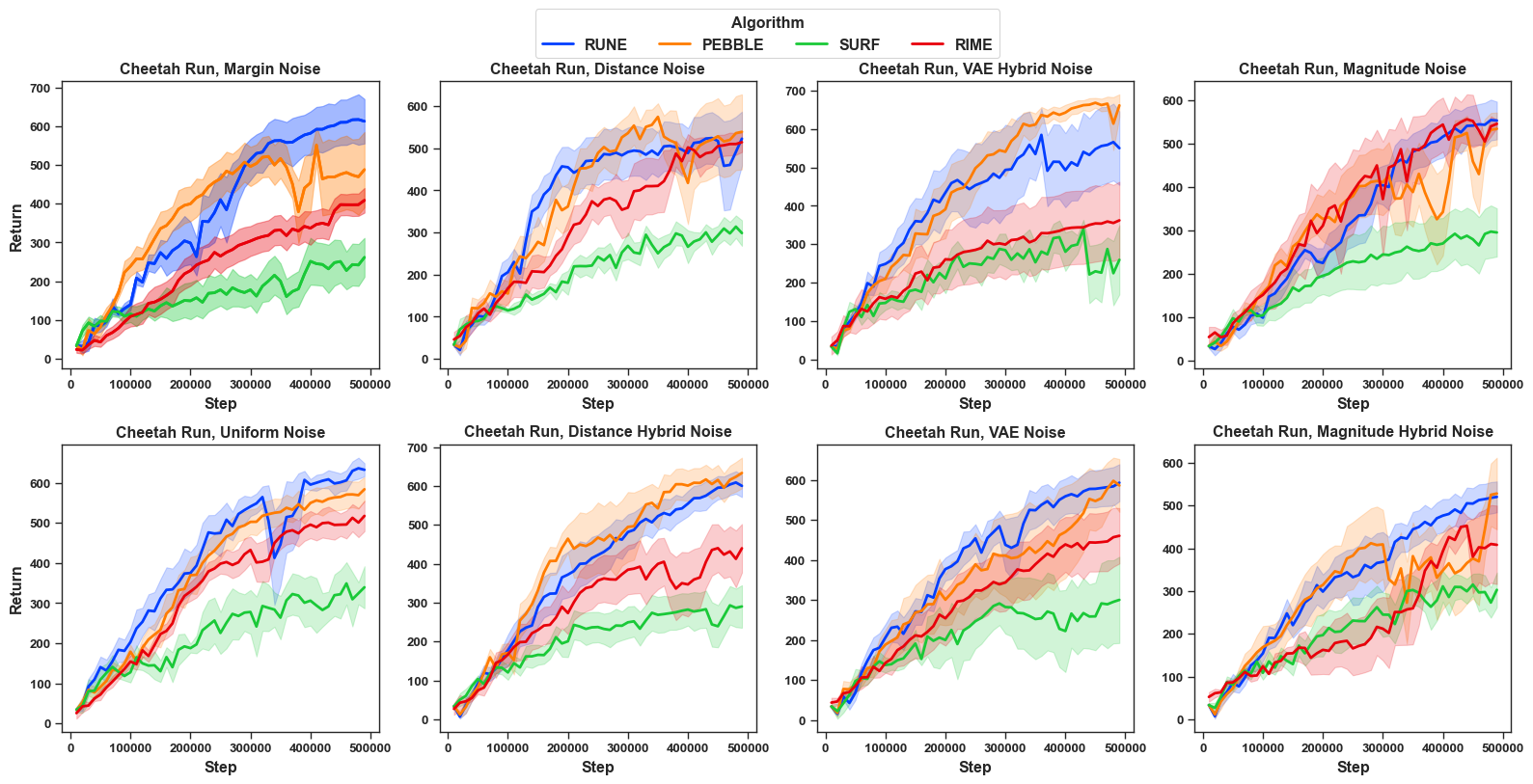}
    \caption{Comparison over different algorithms in 8 types of 20\% noise, in Cheetah Run.}
    \label{fig:compare_algo_0.2}
\end{figure*}

\subsection{Adversarial Noise Study}
To evaluate robustness under challenging corruption, we also construct a model-consistent adversarial noise process inspired by the denoising criterion used in RIME~\citep{Cheng2024RIME}. RIME relies on KL divergence to detect noisy labels, where the KL divergence between noisy preference and predicted logits from the reward model is higher.
We inject noise into samples that gives a low KL divergence between the prediction from the reward function and the incorrect preference that's opposite to the ground truth. In other words, the noise is injected into labels and trajectories where it's most likely to bypass RIME's denoise mechanism by having a small KL divergence with false labels, as shown in Equation~\ref{equ:adversarial}. Here, $T_{\theta_t}(\tau_1,\tau_2)$ is the distribution of preferring each trajectory, given the current learnt reward function under the Bradley-Terry model, and $T_w(\tau_1,\tau_2))$ is the wrong teacher, which will always give the opposite prediction to the oracle teacher $T_o$. The threshold $t$ is similarly determined by the top $\epsilon\%$ KL divergence candidates.
Unlike margin dependent Noise, this type of noise is hypothetical, as it requires access to ground truth labels.
\begin{equation}
    N(\tau_1, \tau_2) = 
\begin{cases}
1 & \text{if } Div(T_{\theta_t}(\tau_1,\tau_2)||T_w(\tau_1,\tau_2)) < t, \\
0  & \text{if } Div(T_{\theta_t}(\tau_1,\tau_2)||T_w(\tau_1,\tau_2)) \ge t.
\end{cases}
\label{equ:adversarial}
\end{equation}

While adversarial noise is injected to attack the KL-divergence-based denoising techniques with the knowledge of the ground truth, it is found that this method, surprisingly, does not always work. The results are shown in Figure~\ref{fig:comapre_noise}, denoted as \textbf{Adversarial} (yellow). For example, we see that in Walker, adversarial noise constantly gives a higher episodic return than Uniform noise, while in other domains, adversarial noise is generally much harder than Uniform Noise. This pattern is consistent with Margin Dependent Noise's results, and it suggests that the current noise-robust PbRL methods can have domain dependency in denoising ability. Here to answer our research questions, the adversarial noise shows a similar pattern\footnote{Their influence towards episodic return in comparison with uniform noise shows moderate positive correlation with a Pearson's Correlation Coefficient of 0.57.} with margin dependent noise and is generally harder than uniform noise.

%% file: chapters/related_work.tex
\section{Related work}

\noindent \textbf{Preference-based Reinforcement Learning:} The motivation behind PbRL is that reward functions are often manually engineered by trial and error and not correlated to an actual task metric~\citep{knox2023reward,booth2023perils}. Hence, 
this paradigm does not require access to a reward function. Instead, a reward function is learned from comparative feedback called preference over pairs of trajectories~\citep{christiano2017deep} from humans using the Bradley-Terry model.
A recent notable success is LLM fine-tuning~\citep{ouyang2022training,rafailov2023direct}, to align the LLM responses in accordance with human preference.
A state-of-the-art algorithm, PEBBLE improves sample efficiency of PbRL by introducing unsupervised pretraining~\citep{lee2021pebble} followed by recent advances ~\citep{DBLP:conf/iclr/ParkSSLAL22,DBLP:conf/iclr/LiangSLA22,feng2025duo} with respect to preference annotation, query diversity, and sampling strategies. \\
\noindent \textbf{Teacher models in PbRL:} Most of the above-mentioned prior work, including PEBBLE, assumes preferences from a perfectly scripted teacher, which is not ideal. To alleviate this assumption, Lee at al.~\citep{lee2021bpref} proposed several models of simulating actual human behavior, including mistakes and myopic scripted teachers. Moreover, they introduced an equally preferable teacher with preferences sampled from a uniform distribution $(0.5,0.5)$ if the two trajectory pairs are value-wise similar. They also observed that a noise level of as small as $10\%$ led to poor performance of the agent. Our work is inspired by Lee et al.'s work on proposing realistic models of irrational teachers. However, we go beyond these simple models and formalize complex noise functions to model teacher errors within PbRL.
There is also prior work that uses LLM/VLM as a teacher to provide preference ~\citep{li2025well,wang2024rl,tu2024online}. However, LLM/VLM preferences rely on strong models like GPT, and its noise's influence on policy learning within PbRL has not been explored.
\\
\noindent \textbf{Noise Robust Techniques in PbRL:} In supervised learning literature, identifying, filtering, and correcting noisy labels have been widely studied with techniques like the small-loss trick~\citep{DBLP:conf/iclr/ZhangBHRV17,younesian2021qactor}, co-teaching among peer networks~\citep{han2018co} and learning the noise transition matrix~\citep{patrini2017making}. \citep{DBLP:conf/ijcai/Xue0YX24} learned a reward function from inconsistent and diverse annotators by using an encoder-decoder-based architecture in latent space and computing reward uncertainty in that space. However, they used the stochastic teacher model from Lee et al.'s work and focused on diverse annotators but not on robustness against noise. Adapting the idea of the small-loss trick, RIME~\citep{Cheng2024RIME} proposed a denoising discriminator mechanism where the trustworthy preference sample is identified as the ones with low KL-divergence between the observed and the predicted preference, along with correction of noisy samples by flipping their labels. They achieve superior agent performance with a noise level of up to $30\%$. In another recent work~\cite{huang2025trend}, adapted co-teaching from supervised learning literature between an ensemble of three reward models to teach each other using identified clean samples showing robustness against noise upto $40\%$; however, they had to utilize demonstrations to mitigate the effects of noise.
All of the previously mentioned noise-robust methods show good performance with the uniform noise model; i.e., with a fixed probability, preference labels are flipped in these settings. Though the idea of feature-dependent noise exists within supervised literature~\citep{yao2021instance, ouyang2022training, xia2020part, DBLP:conf/iclr/ZhangZW0021},  to the best of our knowledge, we are the first to introduce the idea of \textit{feature-dependent noise within the PbRL framework}.

%% file: chapters/conclusion.tex
\section{Conclusion}
This work introduced models of irrational teachers within the Preference-based Reinforcement Learning (PbRL) framework by formalizing feature-dependent noise, where a teacher’s feedback depends on specific trajectory features. We proposed several such noise types—feature magnitude, feature similarity, and margin dependent noise—and evaluated them using a state-of-the-art denoising algorithm designed for uniform noise. Our results show that feature-dependent noise can be harder to detect due to its correlation with underlying features, highlighting the need for methods that can identify structured noise. Future work will explore denoising algorithms tailored to such noise and user studies to understand how often non-experts induce these biases.

%% file: chapters/appendix.tex
\section{Appendix}

\label{app:exp_settings}

\subsection{Implementation details}
We adopted RIME\cite{Cheng2024RIME} as our test bed. Each environment is initialized with its corresponding MuJoCo configuration. The training system performs alternating operations between reward model updates and policy optimization.
The replay buffer receives new labels from reward updates, which maintain the learned reward function in alignment with policy actions.

The agent uses intrinsic state-entropy rewards to build up the replay buffer during the unsupervised pre-training phase (\texttt{unsup\_steps}) before the teacher preferences become available. The system selects feedback samples through adaptive methods based on the chosen feed type, which includes uniform, disagreement, entropy, or $k$-center, until it exhausts the maximum feedback budget.

All experiments were executed on NVIDIA A40/L40S GPUs with CUDA acceleration. Each run was repeated across 5 random seeds for statistical stability, and the reported results correspond to the mean and standard deviation across seeds.

\subsection{Experimental Settings of Hybrid Noise}

For all experiments, we follow the RIME framework settings with task-specific adjustments to stabilize training across different domains. The number of unsupervised steps (\texttt{unsup\_steps}) varies slightly by environment, while other components remain constant.
\begin{table}[h]
\centering
\scriptsize
\renewcommand{\arraystretch}{0.9}
\setlength{\tabcolsep}{3pt}
\resizebox{\columnwidth}{!}{
\begin{tabular}{lccccc}
\toprule
\textbf{Environment} & \textbf{Unsupervised Steps} & \textbf{SAC LR} & \textbf{Interactions} & \textbf{Feedback} & \textbf{Reward Batch} \\
\midrule
Walker-Walk & 9000 & 5e-4 & 20,000 & 1,000 & 100 \\
Cheetah-Run & 2000 & 5e-4 & 20,000 & 1,000 & 100 \\
Quadruped-Walk & 9000 & 1e-4 & 30,000 & 4,000 & 400 \\
MetaWorld Button-Press-V2 & 9000 & 3e-4 & 5,000 & 20,000 & 100 \\
MetaWorld Sweep-Into-V2 & 9000 & 3e-4 & 5,000 & 20,000 & 100 \\
MetaWorld Hammer-V2 & 9000 & 3e-4 & 5,000 & 80,000 & 400 \\
\bottomrule
\end{tabular}
}
\caption{\small Environment-specific hyperparameters used in RIME across DMControl and MetaWorld tasks.}
\label{tab:rime_exp_settings}
\end{table}

Table ~\ref{tab:alpha_values} contains the set of $\alpha$ values that have been pointed out as the optimum values for all experiments of Magnitude Hybrid Noise. \\
For Similarity Hybrid Noise, the optimum $\alpha$ value was 0.5 for all scales of noise, indicating equal weighting in the scores of Margin Dependent Noise and Trajectory Similarity Noise.

\begin{table}[t]
\centering
\scriptsize
\renewcommand{\arraystretch}{0.9}
\setlength{\tabcolsep}{6pt}
\begin{tabular}{lcccc}
\toprule
\textbf{DMControl Domain} & \textbf{10\%} & \textbf{20\%} & \textbf{30\%} & \textbf{40\%} \\
\midrule
Walker Walk & 0.5 & 0.5 & 0.7 & 0.9\\
HalfCheetah Run & 0.9 & 0.3 & 0.3 & 0.1 \\
Quadruped Walk  & 0.9 & 0.5 & 0.5 & 0.3\\
\bottomrule
\end{tabular}
\caption{\small Optimum $\alpha$ values found for Magnitude Hybrid Noise after experimenting multiple cases with $\alpha \in [0, 1]$.}
\label{tab:alpha_values}
\end{table}

\subsection{VAE Encoder Settings}
We use VAE encoders in our VAE Encoding Distance Noise experiments. These encoders are trained on the collected trajectories on a previous normal run of Preference-Based Reinforcement Learning. 
For Cheetah, we train our encoders on an MLP neural network.
For Quadruoped and Walker, where the observation dimension is higher and requires a stronger encoder, we choose a transformer structure. The hyperparameters are shown in Table~\ref{tab:vae_settings}.

\begin{table}[htbp]
    \centering
    \scriptsize
    \resizebox{\columnwidth}{!}{
    \begin{tabular}{cccc}
         & Cheetah & Walker & Quadruped\\
         \midrule
        Structure & MLP Only & Transformer & Transformer\\
        Learning rate & 1e-4 & 1e-4 & 1e-4\\
        Epochs & 1e5 & 1e5 & 1e5\\
        Batch Size & 128 & 128 & 128\\
        Reconstruction Loss Weight & 1 & 1 & 1\\
        KL Loss Weight & 1 & 1 & 1\\
        Input Size & 1150 & 1500 & 4500 \\
        Embedding Size & 128 & 256 & 512 \\
        Encoder Hidden Sizes & 1024-512-256 & N/A & N/A \\
        Transformer Layers & N/A & 2 & 2 \\
        Transformer Heads & N/A & 4 & 4 \\
        Transformer Dropout & N/A & 0.0 & 0.0 \\
         \bottomrule
    \end{tabular}
    }
    \caption{\small Hyperparameters for VAE encoder training.}
    \label{tab:vae_settings}
\end{table}

\section{Results on More Domains}
\label{app:meta_world}
\begin{figure*}[h]
    \centering
    \includegraphics[width=1\linewidth]{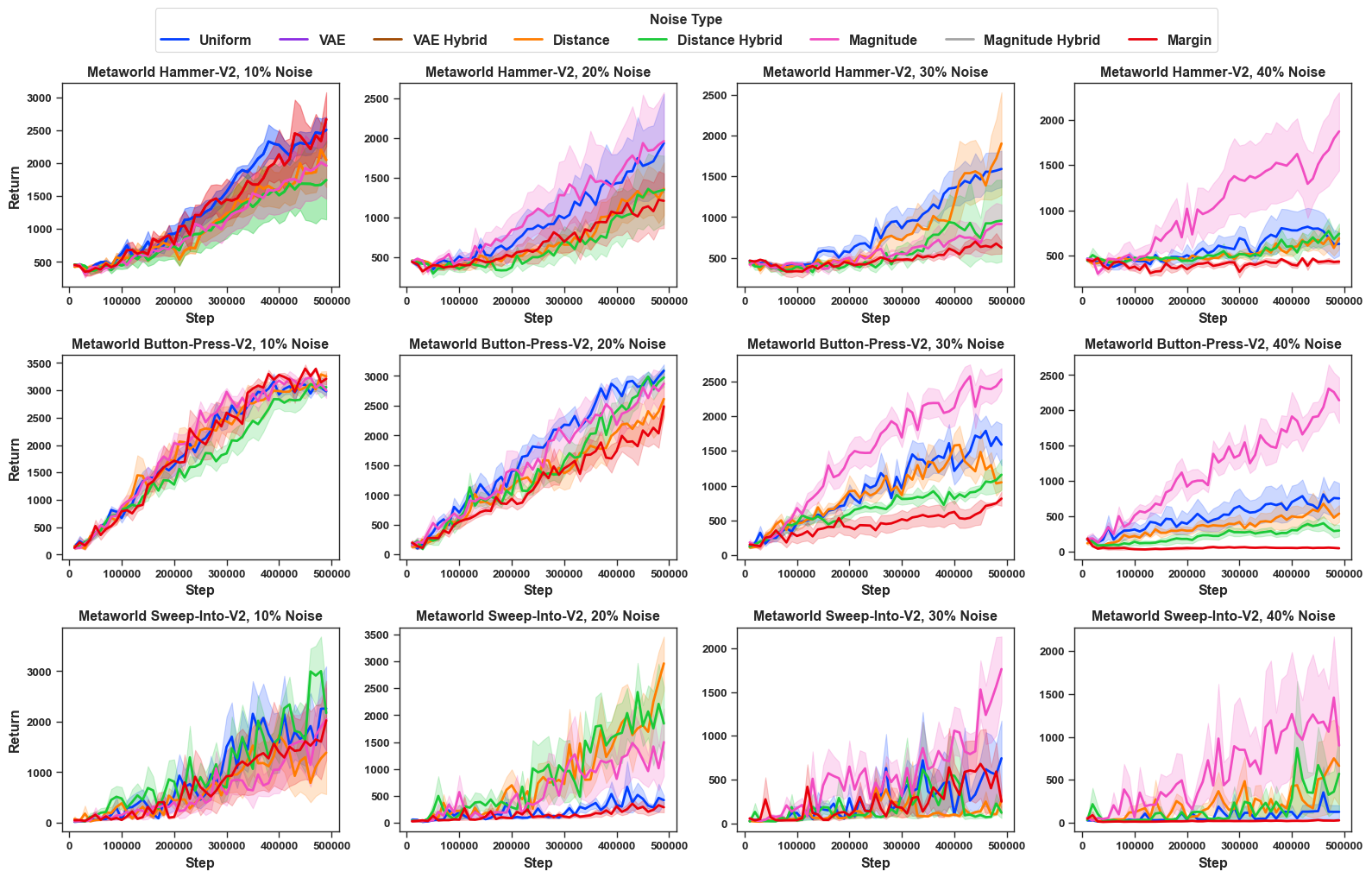}
    \caption{Results on Metaworld Domains. The x-axis is training steps, and     the y-axis is episodic return.}
    \label{fig:meta_world_compare}
\end{figure*}

We also show our evaluation results on three Metaworld domains: Metaworld Button Press, Metaworld Sweep Into and Metaworld Hammer. The results are shown in Figure~\ref{fig:meta_world_compare}. While exceptions exist, we see margin dependent noise, adversarial noise, and hybrid noise with L2 distance are often harder than uniform noise, and trajectory feature magnitude noise is often easier. This is consistent with our previous results in DMControl Domains. \\
\noindent

\emph{Trajectory Feature Noise.} The corruption levels of Metaworld domains produce different results based on the tasks and noise intensity levels. For this noise, we used the displacement of the end-effector as the feature space for all three domains. The results are presented in Figure~\ref{fig:meta_world_compare} as denoted by \textbf{Magnitude} (green). Detailed results on the average final reward and the deviation of the same can be seen in Table~\ref{tab:metaworld_results}. The Hammer results show that Trajectory Feature noise helps because it adds mild variability that improves robustness. The noise level between 20\% and 40\% causes the system to perform actions in an unbalanced manner, which makes it difficult for the agent to determine the superior trajectories. In Button-Press, low noise (10\%) helps, but larger noise corrupts the trajectory labels. Trajectory Feature noise in Sweep-Into at higher levels of noise is relatively easy to handle and shows much higher returns than uniform noise.
\\
\noindent

\emph{Adversarial Noise.} As shown in Figure ~\ref{fig:meta_world_compare} denoted by \textbf{Adversarial(brown)}, the results show that Adversarial noise produces the smallest returns in every domain because it successfully interferes with preference labels. The results from Hammer and Button-Press show that returns decrease sharply when corruption levels exceed 20–30\% because adversarial perturbations create systematic misdirection that causes learning instability. Sweep-Into shows the highest sensitivity because it fails to work with minimal noise levels which means that adversarial perturbations break down preference consistency.\\

\noindent
\emph{Trajectory Similarity Noise.} As shown in Figure ~\ref{fig:meta_world_compare} denoted by \textbf{Distance (red)}, the effect of this noise remains better compared to Uniform noise when the corruption level increases.
The effect of Hammer-V2 becomes more pronounced when noise levels increase.\\

\noindent
\emph{Similarity Hybrid Noise.} As shown in Figure ~\ref{fig:meta_world_compare} denoted by \textbf{Distance Hybrid(purple)}, the degradation pattern of this noise appears more gradual than what occurs with Distance noise or Adversarial noise alone. It performs comparably to Uniform at 10–20\% noise but causes a consistent decline beyond 30\%.\\

\noindent
\emph{Margin Dependent Noise.} As shown in Figure ~\ref{fig:meta_world_compare} denoted by \textbf{Margin(blue)}, it demonstrates stable performance under all noise conditions.
The model's predictive marginal uncertainty leads to a sharp drop in performance between 30\% and 40\% noise.

\begin{table*}[t]
\centering
\renewcommand{\arraystretch}{0.9}
\setlength{\tabcolsep}{3pt}
\scriptsize

\resizebox{0.78\textwidth}{!}{
\begin{tabular}{lcccc}
\toprule
\textbf{Noise Type} & \textbf{10\%} & \textbf{20\%} & \textbf{30\%} & \textbf{40\%} \\
\midrule

\multicolumn{5}{l}{\textbf{Sweep Into}} \\[1pt]
\hspace{6pt} Uniform & 3466.90 $\pm$ 1218.69 & 1330.63 $\pm$ 464.78 & 1717.99 $\pm$ 2078.39 & 219.59 $\pm$ 387.22 \\
\cmidrule(lr){1-5}
\hspace{6pt} Adversarial & \textbf{1416.33 $\pm$ 1826.93} & \textbf{345.57 $\pm$ 639.10} & \textbf{86.54 $\pm$ 42.18} & 39.74 $\pm$ 18.54 \\
\hspace{6pt} Distance & 1586.57 $\pm$ 1650.43 & 2093.10 $\pm$ 1762.54 & 197.49 $\pm$ 86.44 & 403.85 $\pm$ 786.49 \\
\hspace{6pt} Distance Hybrid & 2430.13 $\pm$ 1568.62 & 2324.90 $\pm$ 1246.07 & 384.65 $\pm$ 542.88 & 170.02 $\pm$ 246.53 \\
\hspace{6pt} Magnitude & 2387.58 $\pm$ 2065.93 & 2722.03 $\pm$ 1438.57 & 3896.63 $\pm$ 806.52 & 1792.17 $\pm$ 1322.71 \\
\hspace{6pt} Margin & 2398.88 $\pm$ 2031.31 & 443.84 $\pm$ 516.09 & 1617.74 $\pm$ 977.79 & \textbf{37.13 $\pm$ 20.70} \\

\midrule
\multicolumn{5}{l}{\textbf{Hammer}} \\[1pt]
\hspace{6pt} Uniform & 4057.50 $\pm$ 824.33 & 2615.88 $\pm$ 1586.99 & 2390.72 $\pm$ 855.12 & 972.32 $\pm$ 706.08 \\
\cmidrule(lr){1-5}
\hspace{6pt} Adversarial & 2231.48 $\pm$ 693.40 & \textbf{869.40 $\pm$ 795.95} & \textbf{813.92 $\pm$ 522.78} & \textbf{444.42 $\pm$ 326.89} \\
\hspace{6pt} Distance & \textbf{1377.99 $\pm$ 1175.53} & 1066.00 $\pm$ 747.60 & 851.23 $\pm$ 477.91 & 731.58 $\pm$ 269.28 \\
\hspace{6pt} Distance Hybrid & 1747.67 $\pm$ 1047.95 & 954.41 $\pm$ 766.82 & 820.71 $\pm$ 503.00 & 754.02 $\pm$ 455.96 \\
\hspace{6pt} Magnitude & 3368.95 $\pm$ 1808.27 & 2797.50 $\pm$ 1964.17 & 1476.73 $\pm$ 1028.25 & 2754.59 $\pm$ 906.00 \\
\hspace{6pt} Margin & 2568.43 $\pm$ 883.48 & 1211.88 $\pm$ 847.18 & 859.85 $\pm$ 387.05 & 451.13 $\pm$ 32.21 \\

\midrule
\multicolumn{5}{l}{\textbf{Button Press}} \\[1pt]
\hspace{6pt} Uniform & 3806.30 $\pm$ 114.78 & 3394.65 $\pm$ 324.68 & 2781.93 $\pm$ 498.91 & 1238.34 $\pm$ 933.86 \\
\cmidrule(lr){1-5}
\hspace{6pt} Adversarial & \textbf{3069.07 $\pm$ 195.86} & \textbf{2195.84 $\pm$ 799.51} & \textbf{713.53 $\pm$ 373.31} & 398.31 $\pm$ 611.62 \\
\hspace{6pt} Distance & 3257.28 $\pm$ 170.01 & 2492.25 $\pm$ 837.19& 1306.21 $\pm$ 219.48 & 548.45 $\pm$ 344.40 \\
\hspace{6pt} Distance Hybrid & 3231.84 $\pm$ 177.28 & 3052.83 $\pm$ 236.78 & 1124.13 $\pm$ 325.21 & 362.06 $\pm$ 190.03 \\
\hspace{6pt} Magnitude & 3599.05 $\pm$ 135.07 & 3423.87 $\pm$ 391.25 & 3476.93 $\pm$ 275.48 & 2906.58 $\pm$ 703.96 \\
\hspace{6pt} Margin & 3215.23 $\pm$ 161.07 & 2781.47 $\pm$ 422.75 & 865.07 $\pm$ 604.01 & \textbf{67.82 $\pm$ 9.49} \\

\bottomrule
\end{tabular}}
\caption{Final average return (mean $\pm$ std) across domains (\emph{Sweep Into}, \emph{Hammer}, and \emph{Button Press}) and hue noise levels for each noise type. (--) indicate missing entries.}
\label{tab:metaworld_results}
\end{table*}

\begin{table*}[t]
\centering
\renewcommand{\arraystretch}{0.9}
\scriptsize
\setlength{\tabcolsep}{4pt}

\resizebox{0.38\textwidth}{!}{
\begin{tabular}{lcccc}
\toprule
\textbf{Domain} & \textbf{10\%} & \textbf{20\%} & \textbf{30\%} & \textbf{40\%} \\
\midrule
Walker Walk     & 4 & 1 & 2 & 4 \\
Cheetah Run     & 8 & 6 & 8 & 8 \\
Quadruped Walk  & 8 & 4 & 8 & 4 \\
\midrule
\textbf{Overall (\% FDN $>$ Uniform)} & \textbf{83.3\%} & \textbf{45.8\%} & \textbf{75.0\%} & \textbf{66.7\%} \\
\bottomrule
\end{tabular}}
\caption{\small Number of noise types (out of 8) yielding lower mean return than the \emph{Uniform} baseline for each DMControl domain and noise level. The last row shows the overall percentage of experiments where Feature-Dependent Noise (FDN) outperforms Uniform (out of 24 possible comparisons per noise level).}
\label{tab:harder_than_uniform_dmcontrol}
\end{table*}

\begin{table*}[t]
\centering
\renewcommand{\arraystretch}{0.9}
\scriptsize
\setlength{\tabcolsep}{4pt}

\resizebox{0.38\textwidth}{!}{
\begin{tabular}{lcccc}
\toprule
\textbf{Domain} & \textbf{10\%} & \textbf{20\%} & \textbf{30\%} & \textbf{40\%} \\
\midrule
Sweep Into   & 5 & 2 & 4 & 3 \\
Hammer       & 5 & 3 & 5 & 4 \\
Button Press & 5 & 3 & 4 & 4 \\
\midrule
\textbf{Overall (\% FDN $>$ Uniform)} & \textbf{100.0\%} & \textbf{53.3\%} & \textbf{86.7\%} & \textbf{73.3\%} \\
\bottomrule
\end{tabular}}
\caption{\small Number of noise types (out of 5) yielding lower mean return than the \emph{Uniform} baseline for each Metaworld domain and noise level. The last row reports the percentage of experiments where Feature-Dependent Noise (FDN) outperforms Uniform (out of 15 total comparisons per noise level).}
\label{tab:harder_than_uniform_metaworld}
\end{table*}

\input{chapters/appendix_vlm_part}

\input{chapters/appendix_vae}
\input{chapters/appendix_other_noise}

\input{chapters/appendix_downweight}

\label{app:vae}

%% file: chapters/appendix_vlm_part.tex
\clearpage

\AtBeginEnvironment{tcolorbox}{\nolinenumbers}
\AtEndEnvironment{tcolorbox}{\linenumbers}

\section{VLM Prompt Templates}
\label{app:prompt}
In this section, we present our prompt to elicit preferences. We adapt a similar setting from RL-VLM-F \cite{wang2024rl}: we query VLM to summarise the observations first, and then ask VLM to think about the differences from the image observation summaries: which one is closer to the goal? We refer to these two prompts as the Image Summary Prompt and the Preference Elicitation Prompt.  Furthermore, if the VLM cannot find significant differences between the two images, then we have indifferent preference, and we won't use them in training.

\begin{tcolorbox}[colback=red!5!white,colframe=red!75!black,title=Image Summary Prompt in Cart Pole]
\small
1. What is shown in Image 1?

2. What is shown in Image 2?

3. The goal is to balance the brown pole on the black cart to be upright. Are there any differences between Image 1 and Image 2 in terms of achieving the goal?

<Image 1>

<Image 2>
\end{tcolorbox}

\begin{tcolorbox}[colback=red!5!white,colframe=red!75!black,title= Preference Elicitation Prompt in Cart Pole]
\small
Based on the text below to the questions:

1. What is shown in Image 1?

2. What is shown in Image 2?

3. The goal is to balance the brown pole on the black cart to be upright. Are there any differences between Image 1 and Image 2 in terms of achieving the goal?

<Text Summary of Image Observations>

Is the goal better achieved in Image 1 or Image 2?
Reply with a single line of 0 if Image 1 achieves the goal better, or 1 if Image 2 achieves the goal better.
Reply -1 if unsure or there is no difference.
\end{tcolorbox}

\begin{tcolorbox}[colback=red!5!white,colframe=red!75!black,title=Image Summary Prompt in Metaworld Soccer]
\small
1. What is shown in Image 1?

2. What is shown in Image 2?

3. The goal is to move the soccer ball into the goal. Are there any differences between Image 1 and Image 2 in terms of achieving the goal?

<Image 1>

<Image 2>
\end{tcolorbox}

\begin{tcolorbox}[colback=red!5!white,colframe=red!75!black,title= Preference Elicitation Prompt in Metaworld Soccer]
\small
Based on the text below to the questions:

1. What is shown in Image 1?

2. What is shown in Image 2?

3. The goal is to move the soccer ball into the goal. Are there any differences between Image 1 and Image 2 in terms of achieving the goal?

<Text Summary of Image Observations>

Is the goal better achieved in Image 1 or Image 2?
Reply a single line of 0 if Image 1 achieves the goal better, or 1 if Image 2 achieves the goal better.
Reply -1 if unsure or there is no difference.
\end{tcolorbox}
\section{VLM Noise Examples}
\begin{figure}[htbp]
  \centering
  \begin{subfigure}[]{0.48\columnwidth}
    \includegraphics[width=\linewidth]{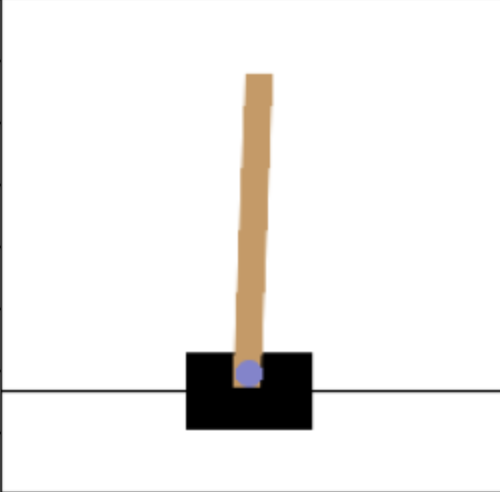}
  \end{subfigure}
  \hfill
  \begin{subfigure}[]{0.48\columnwidth}
    \includegraphics[width=\linewidth]{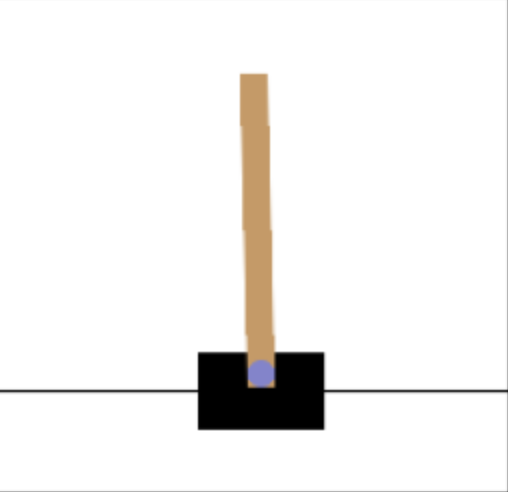}
  \end{subfigure}
  \caption{VLM wrong example, where the two observations are similar.}
  \label{fig:compare_cp_wrong}
\end{figure}

\begin{figure}[htbp]
  \centering
  \begin{subfigure}[]{0.48\columnwidth}
    \includegraphics[width=\linewidth]{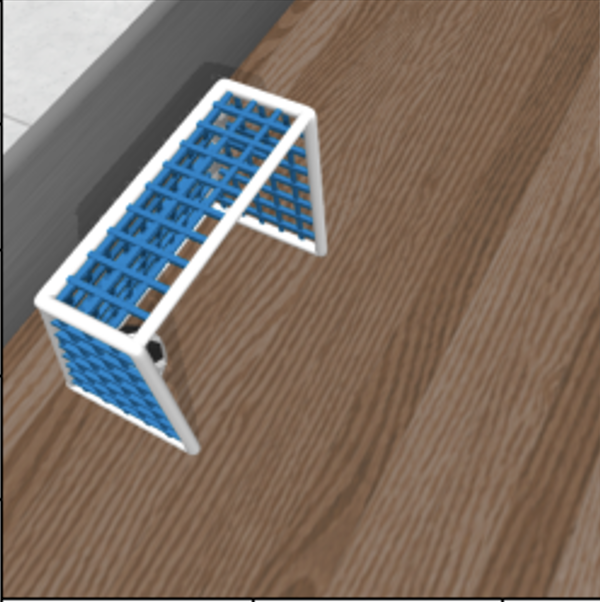}
  \end{subfigure}
  \hfill
  \begin{subfigure}[]{0.48\columnwidth}
    \includegraphics[width=\linewidth]{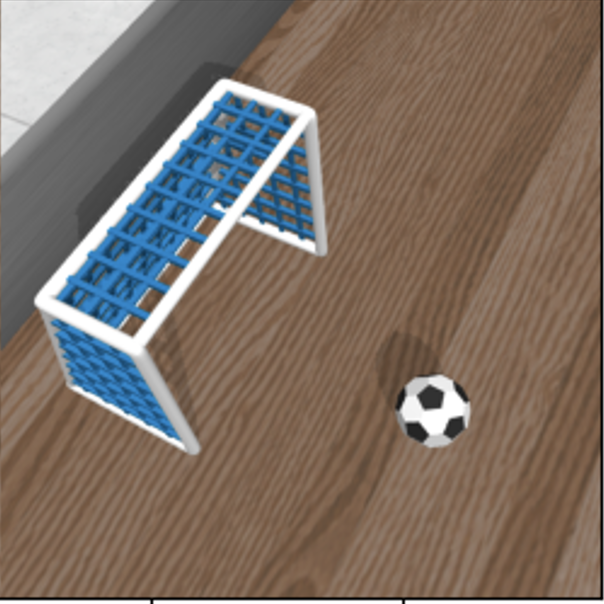}
  \end{subfigure}
  \caption{VLM wrong example, where the left soccer ball is actually already in the goal. This requires detailed observation of the image and goes beyond our VLM's ability.}
  \label{fig:compare_soccer_wrong}
\end{figure}

The VLM gives noisy preferences mostly due to these two reasons: (1) similar observations; (2) it requires image understanding ability beyond the VLM. We present here examples of observations where the VLMs made mistakes in our experiments. These examples are shown in Figure~\ref{fig:compare_cp_wrong} and Figure~\ref{fig:compare_soccer_wrong}. In Figure~\ref{fig:compare_cp_wrong}, while the left image and right images show rods leaning towards  right and left, the angles are very similar, and the VLM cannot give the correct preferences. 
In Figure~\ref{fig:compare_soccer_wrong}, in the left image, the soccer is actually already in the goal, while the VLM did not notice and wrongly interpreted the image as "the soccer ball is outside of view", hence giving incorrect preference.

%% file: chapters/appendix_other_noise.tex
\section{Results on other algorithms}
\label{app:other_algo}
\begin{figure*}[!t]
    \centering
    \includegraphics[width=1\linewidth]{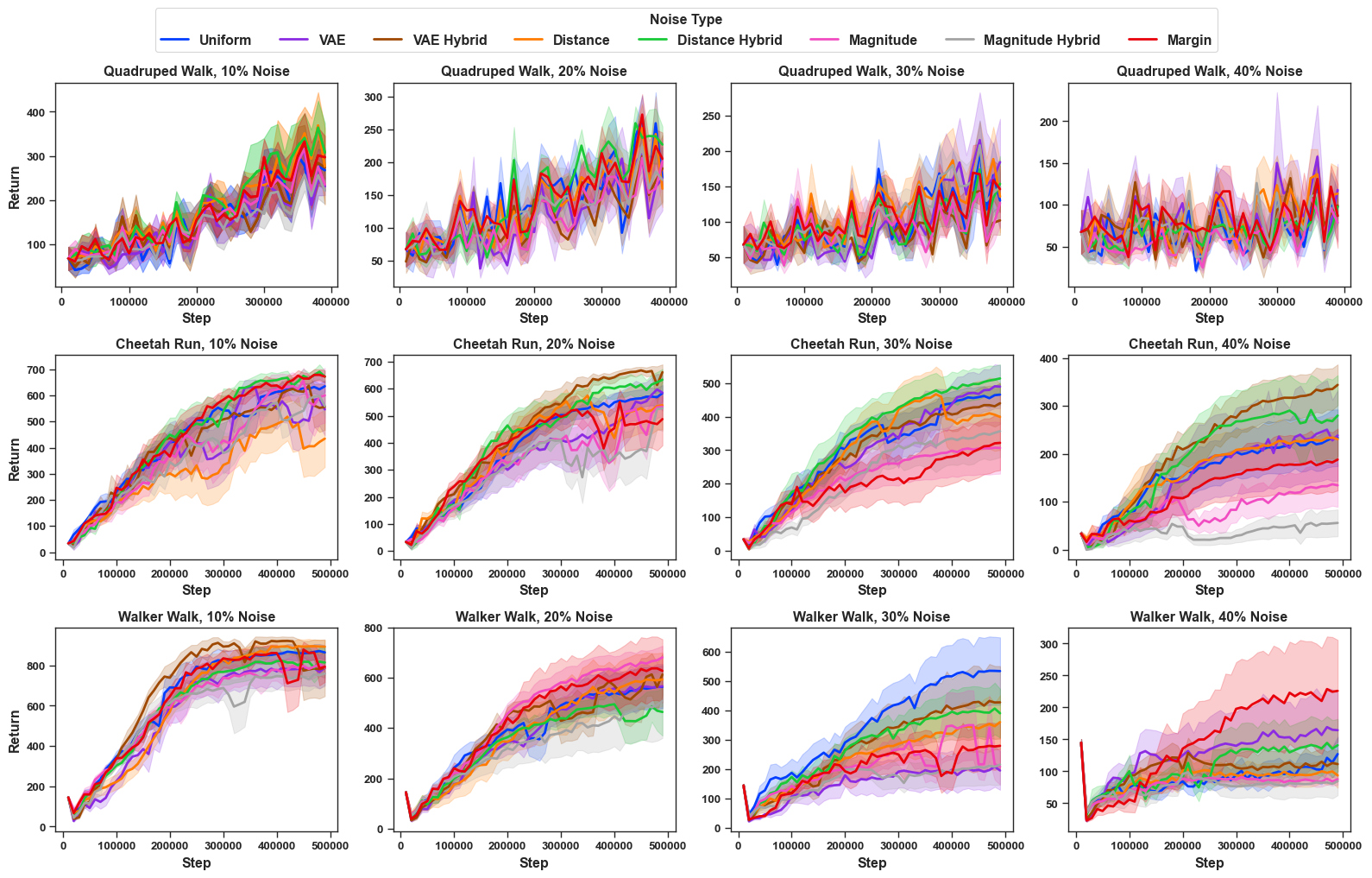}
    \caption{Results on PEBBLE in different proportions and types of noise.}
    \label{fig:comapre_pebble}
\end{figure*}

\begin{figure*}[!t]
    \centering
    \includegraphics[width=1\linewidth]{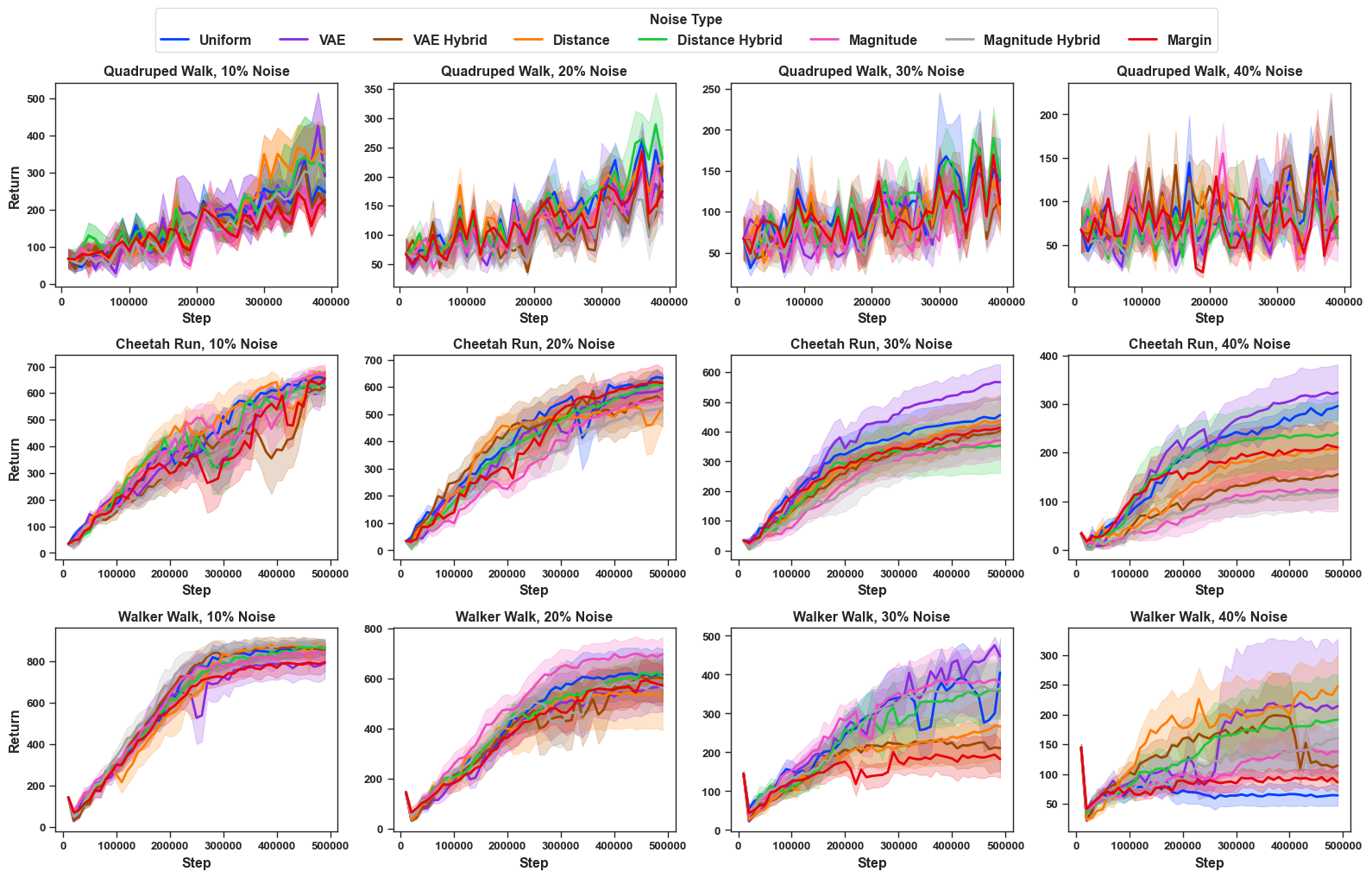}
    \caption{Results on RUNE in different proportions and types of noise.}
    \label{fig:comapre_rune}
\end{figure*}

\begin{figure*}[!t]
    \centering
    \includegraphics[width=1\linewidth]{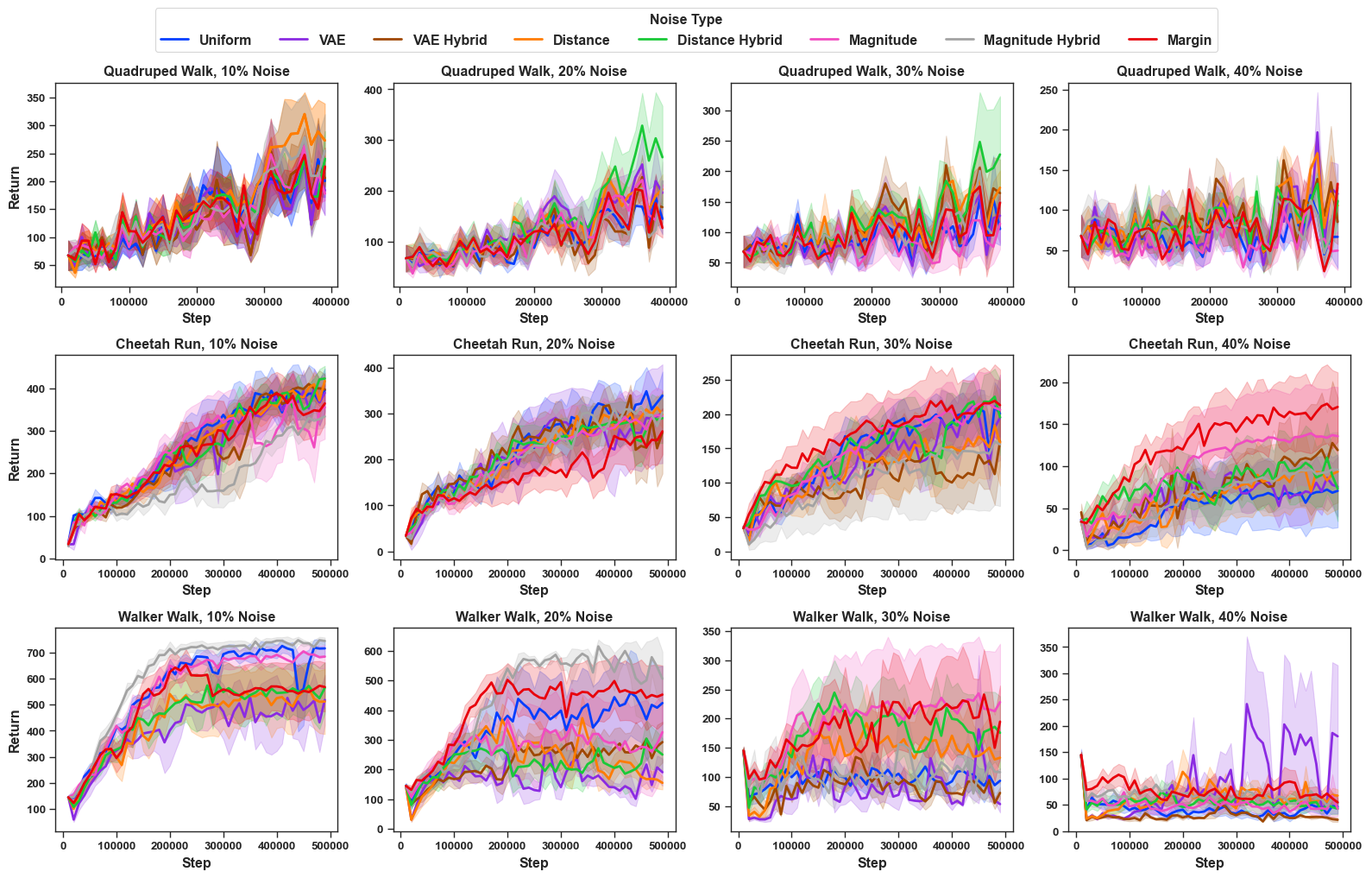}
    \caption{Results on SURF in different proportions and types of noise. }
    \label{fig:comapre_surf}
\end{figure*}

While RIME~\cite{Cheng2024RIME} is one state-of-the-art de-noising PbRL algorithm, we also benchmarked the learning performance of other PbRL algorithms that do not explicitly handle noisy preference, including PEBBLE~\cite{lee2021pebble}, SURF~\cite{DBLP:conf/iclr/ParkSSLAL22} and RUNE~\cite{liang2022reward}. The results are shown in Figure~\ref{fig:comapre_pebble}, Figure~\ref{fig:comapre_rune} and Figure~\ref{fig:comapre_surf} respectively. The corresponding final episodic return are shown in Table~\ref{tab: episodic_return_all_noise_pebble}, Table~\ref{tab: episodic_return_all_noise_rune}, Table~\ref{tab: episodic_return_all_noise_surf}.
Across methods, we observe a similar qualitative pattern to RIME: different noise types induce markedly different levels of difficulty. For example, under RUNE, margin noise and magnitude-hybrid noise are generally more challenging as compared uniform noise. However, these trends these trends do not hold consistently across all algorithms and vary with the underlying algorithm. In PEBBLE, margin noise sometimes leads to substantially lower return (more influence on algorithm) than uniform noise usually for higher scales of noise like 30\% and 40\% noise, but in other cases the ordering reverses. Still, magnitude-hybrid noise is consistently harder than uniform noise for PEBBLE (91\% cases in our experiments across domains and scales of noise). For the remaining noise types, performance differences are often irregular and non-monotonic, suggesting that algorithms without explicit denoising mechanisms can be fragile under noisy preference supervision induced by FDN.

\begin{figure*}[!t]
    \centering
    \includegraphics[width=1\linewidth]{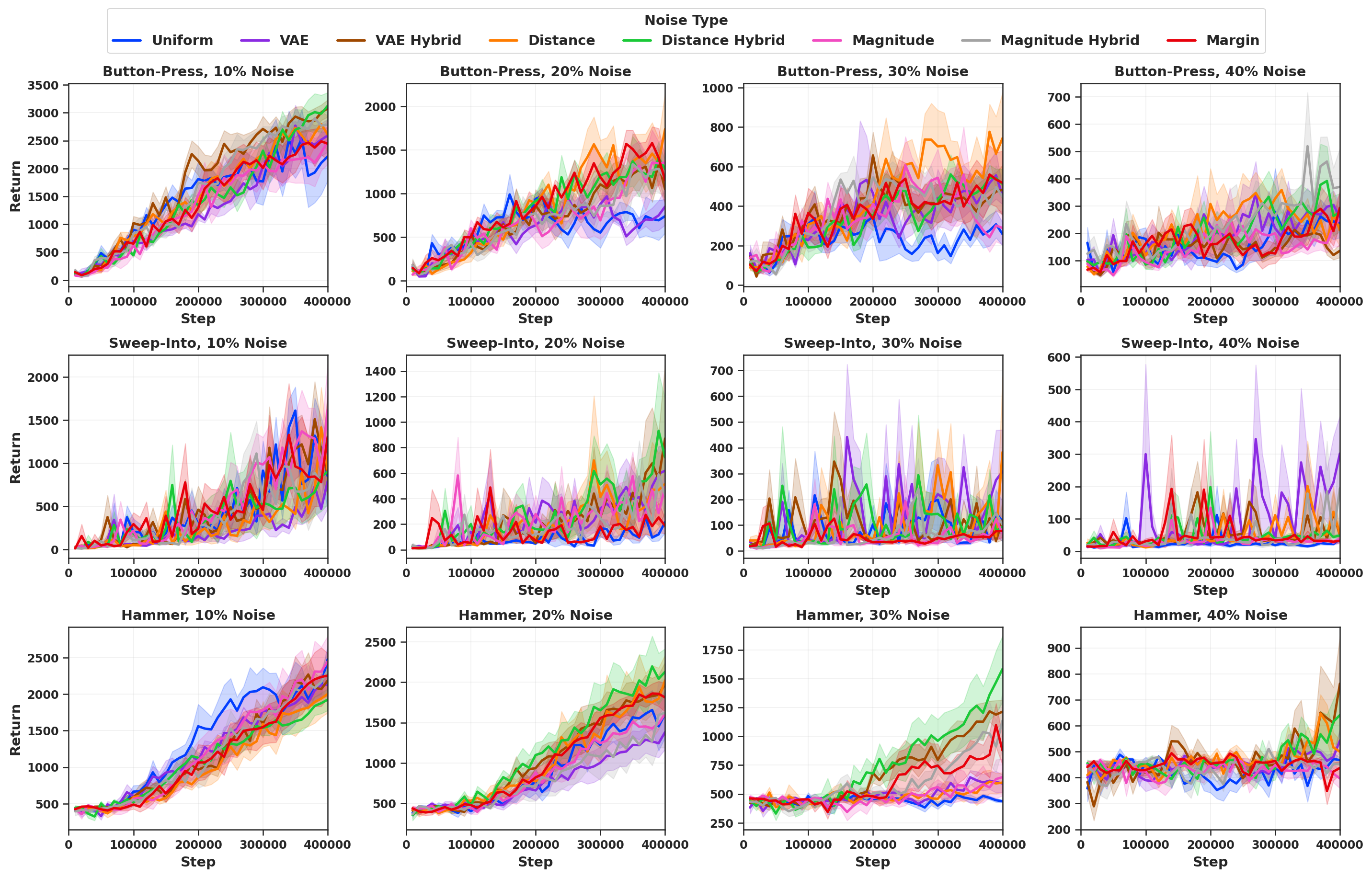}
    \caption{Results on PEBBLE in different proportions and types of noise in Metaworld.}
    \label{fig:comapre_pebble_metaworld}
\end{figure*}

\begin{figure*}[!t]
    \centering
    \includegraphics[width=1\linewidth]{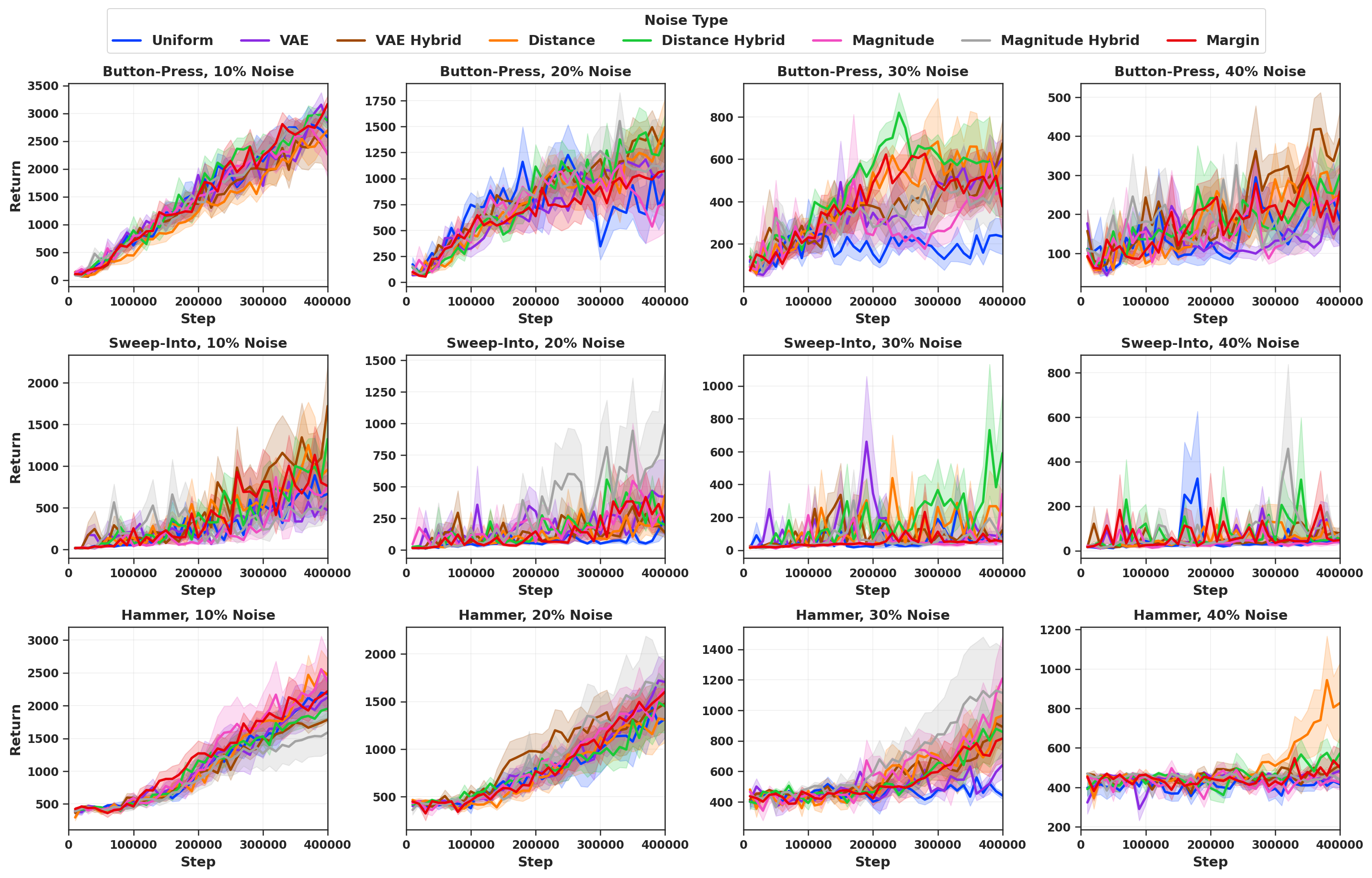}
    \caption{Results on RUNE in different proportions and types of noise in Metaworld.}
    \label{fig:comapre_rune_metaworld}
\end{figure*}

\begin{figure*}[!t]
    \centering
    \includegraphics[width=1\linewidth]{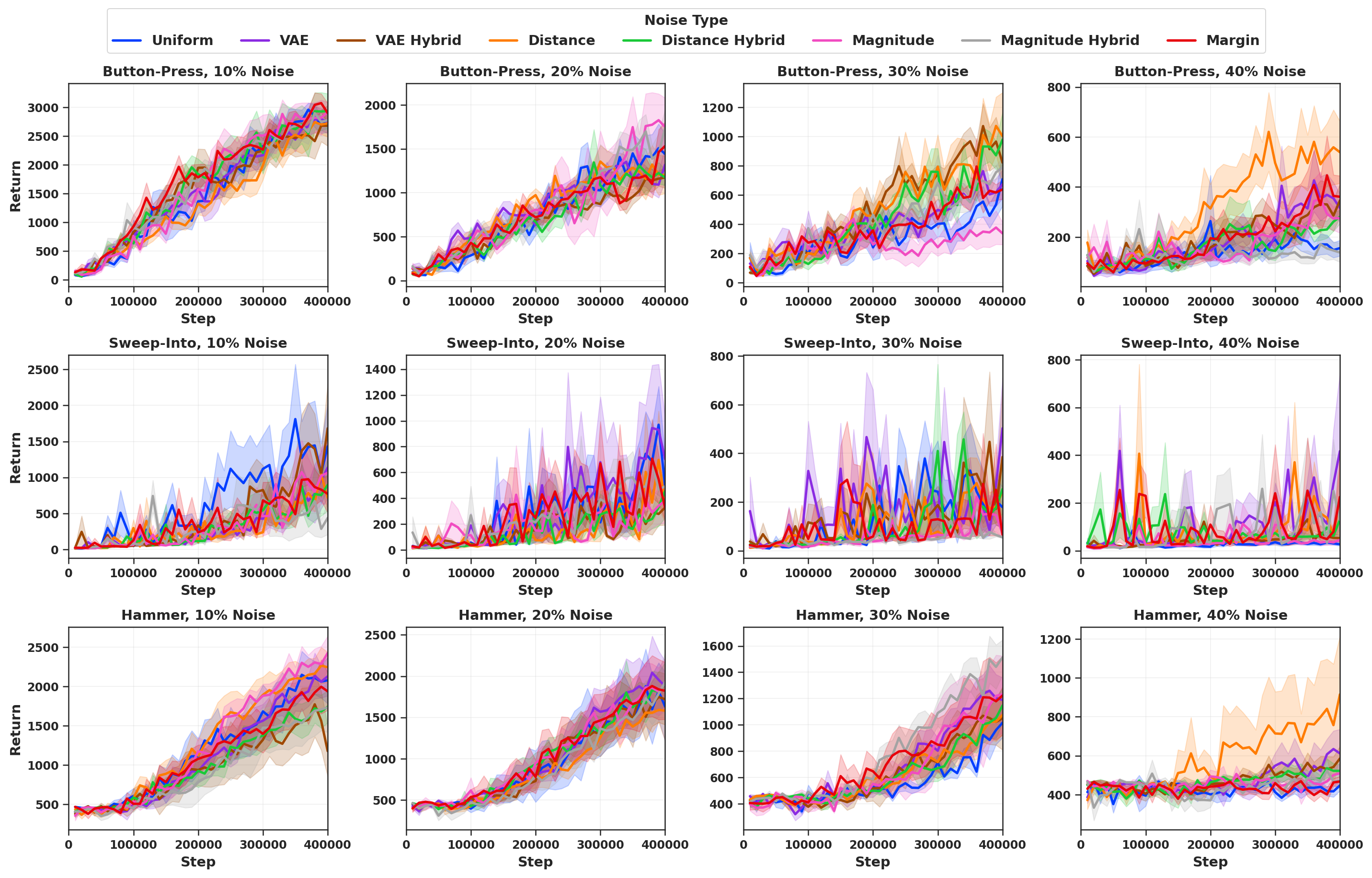}
    \caption{Results on SURF in different proportions and types of noise in Metaworld.}
    \label{fig:comapre_surf_metaworld}
\end{figure*}

We also show results in in Metaworld series domains are shown in Figure~\ref{fig:comapre_pebble_metaworld}, Figure~\ref{fig:comapre_rune_metaworld}, Figure~\ref{fig:comapre_surf_metaworld} and Table~\ref{tab:metaworld_results_pebble}, Table~\ref{tab:metaworld_results_rune}, Table~\ref{tab:metaworld_results_surf}. While these domains are generally harder with more complicated action and observation space, we notice that VAE Hybrid are often more difficult to deal with in comparison with other types of noise. Furthermore, uniform noise here is also very challenging across many settings, especially in Button Press. Again, these results suggest that the learning dynamics within FDN in different algorithms are very domain-dependent and dealing with uniform noise does not necessarily ensure desirable performance in other types of noise. 

\begin{figure*}[!t]
    \centering
    \includegraphics[width=1\linewidth]{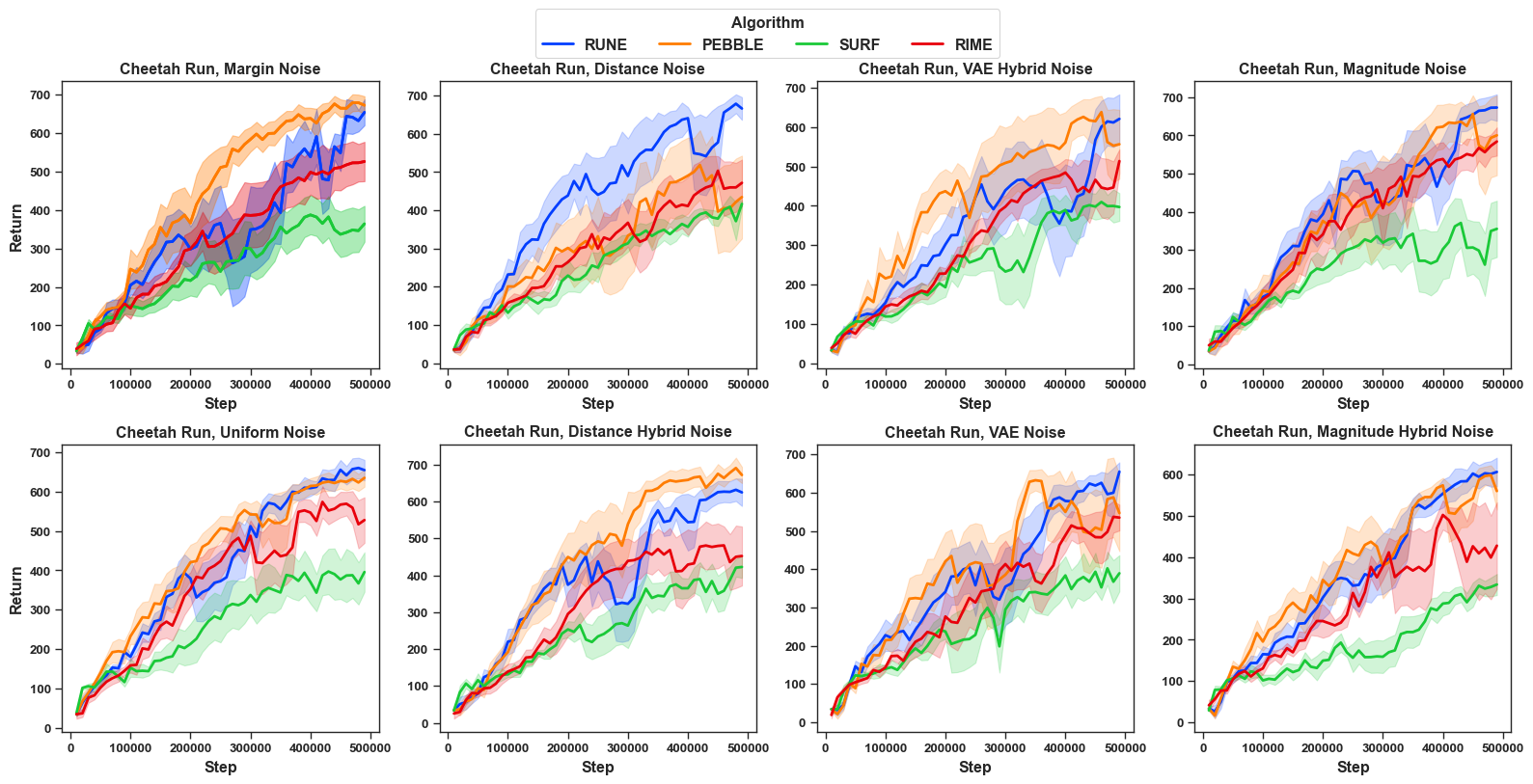}
    \caption{Comparison over different algorithms in 8 types of 10\% noise, in Cheetah Run.}
    \label{fig:compare_algo_0.1}
\end{figure*}

\begin{figure*}[!t]
    \centering
    \includegraphics[width=1\linewidth]{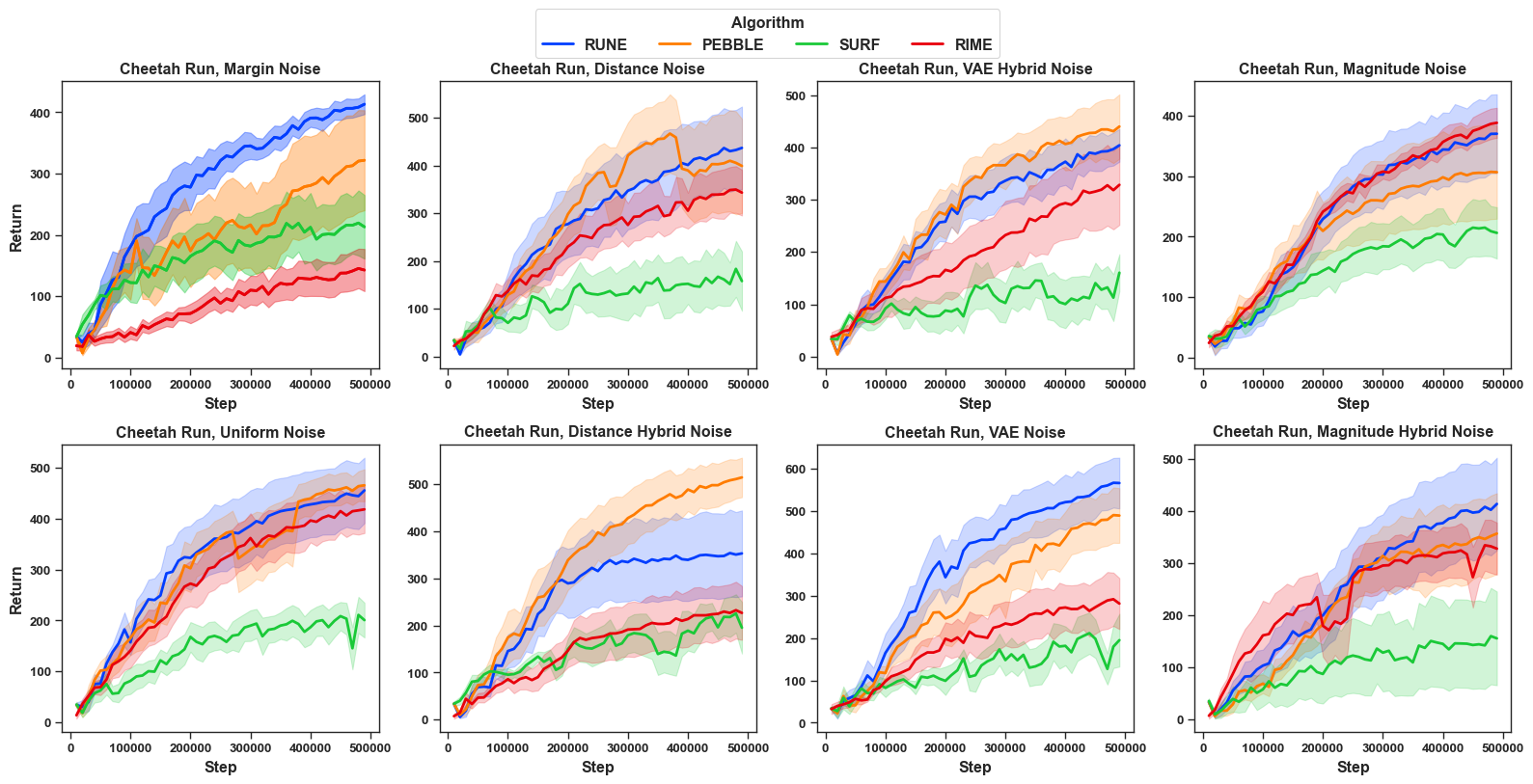}
    \caption{Comparison over different algorithms in 8 types 30\% noise, in Cheetah Run.}
    \label{fig:compare_algo_0.3}
\end{figure*}
\begin{figure*}[!t]
    \centering
    \includegraphics[width=1\linewidth]{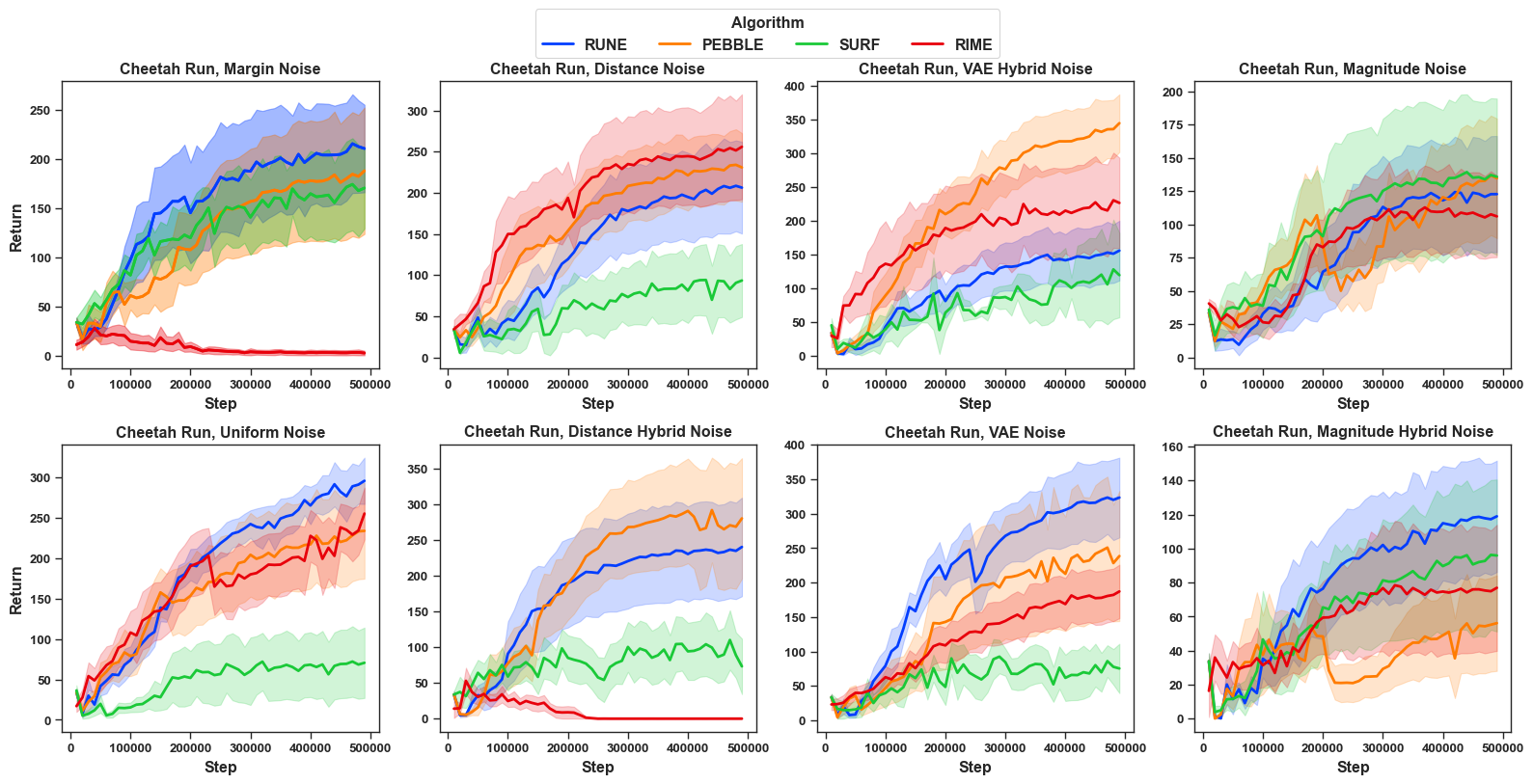}
    \caption{Comparison over different algorithms in 8 types 40\% noise, in Cheetah Run.}
    \label{fig:compare_algo_0.4}
\end{figure*}

\input{chapters/tables/pebble_table}

\input{chapters/tables/rune_table}

\input{chapters/tables/surf_table}

%% file: chapters/tables/pebble_table.tex
\begin{table*}[t]
\centering
\label{tab:pebble_results}

\renewcommand{\arraystretch}{0.9}
\setlength{\tabcolsep}{3pt}
\scriptsize

\resizebox{0.78\textwidth}{!}{
\begin{tabular}{lcccc}
\toprule
\textbf{Noise Type} & \textbf{10\%} & \textbf{20\%} & \textbf{30\%} & \textbf{40\%} \\
\midrule

\multicolumn{5}{l}{\textbf{Walker Walk}} \\[1pt]
\hspace{6pt} Uniform & 864.22 $\pm$ 87.11 & 563.25 $\pm$ 207.64 & 534.75 $\pm$ 281.17 & 127.00 $\pm$ 66.21 \\
\cmidrule(lr){1-5}
\hspace{6pt} Distance & 892.55 $\pm$ 80.98 & 593.34 $\pm$ 262.22 & \textbf{360.81 $\pm$ 224.36} & \textbf{93.58 $\pm$ 47.79} \\
\hspace{6pt} Distance Hybrid & \textbf{814.50 $\pm$ 180.30} & \textbf{464.09 $\pm$ 230.24} & \textbf{391.01 $\pm$ 203.36} & 140.82 $\pm$ 100.46 \\
\hspace{6pt} Magnitude & \textbf{788.76 $\pm$ 135.95} & 680.79 $\pm$ 113.16 & \textbf{208.98 $\pm$ 149.93} & \textbf{87.51 $\pm$ 21.40} \\
\hspace{6pt} Magnitude Hybrid & \textbf{759.51 $\pm$ 154.18} & \textbf{507.94 $\pm$ 361.33} & \textbf{216.27 $\pm$ 167.80} & \textbf{84.56 $\pm$ 54.50} \\
\hspace{6pt} Margin & \textbf{793.52 $\pm$ 186.17} & 627.64 $\pm$ 306.72 & \textbf{280.02 $\pm$ 188.83} & 225.93 $\pm$ 195.45 \\
\hspace{6pt} VAE & \textbf{787.68 $\pm$ 210.09} & 566.21 $\pm$ 215.26 & \textbf{196.04 $\pm$ 161.74} & 164.36 $\pm$ 143.87 \\
\hspace{6pt} VAE Hybrid & \textbf{787.79 $\pm$ 344.68} & 612.62 $\pm$ 201.05 & \textbf{428.20 $\pm$ 302.68} & \textbf{111.28 $\pm$ 54.32} \\
\midrule
\multicolumn{5}{l}{\textbf{HalfCheetah Run}} \\[1pt]
\hspace{6pt} Uniform & 635.41 $\pm$ 55.39 & 583.87 $\pm$ 73.94 & 465.79 $\pm$ 77.31 & 233.84 $\pm$ 145.04 \\
\cmidrule(lr){1-5}
\hspace{6pt} Distance & \textbf{434.70 $\pm$ 265.67} & \textbf{539.14 $\pm$ 218.43} & \textbf{399.31 $\pm$ 242.04} & \textbf{230.95 $\pm$ 102.26} \\
\hspace{6pt} Distance Hybrid & 672.07 $\pm$ 54.13 & 633.65 $\pm$ 96.46 & 514.11 $\pm$ 102.73 & 280.14 $\pm$ 206.36 \\
\hspace{6pt} Magnitude & \textbf{600.25 $\pm$ 256.52} & \textbf{533.64 $\pm$ 92.18} & \textbf{306.57 $\pm$ 188.31} & \textbf{134.63 $\pm$ 110.04} \\
\hspace{6pt} Magnitude Hybrid & \textbf{559.40 $\pm$ 109.29} & \textbf{528.09 $\pm$ 205.35} & \textbf{356.47 $\pm$ 189.67} & \textbf{56.16 $\pm$ 68.72} \\
\hspace{6pt} Margin & 672.50 $\pm$ 57.78 & \textbf{488.00 $\pm$ 237.94} & \textbf{321.75 $\pm$ 200.11} & \textbf{188.27 $\pm$ 157.80} \\
\hspace{6pt} VAE & \textbf{546.63 $\pm$ 244.52} & 585.73 $\pm$ 162.02 & 489.65 $\pm$ 158.39 & 238.60 $\pm$ 231.34 \\
\hspace{6pt} VAE Hybrid & \textbf{556.07 $\pm$ 214.06} & 661.35 $\pm$ 70.78 & \textbf{440.03 $\pm$ 150.95} & 344.16 $\pm$ 104.71 \\
\midrule
\multicolumn{5}{l}{\textbf{Quadruped Walk}} \\[1pt]
\hspace{6pt} Uniform & 267.60 $\pm$ 92.96 & 175.49 $\pm$ 31.30 & 130.07 $\pm$ 78.09 & 97.99 $\pm$ 42.20 \\
\cmidrule(lr){1-5}
\hspace{6pt} Distance & 276.32 $\pm$ 193.05 & \textbf{160.11 $\pm$ 63.19} & 135.65 $\pm$ 113.36 & 113.42 $\pm$ 70.87 \\
\hspace{6pt} Distance Hybrid & 307.56 $\pm$ 148.98 & 227.41 $\pm$ 55.68 & 153.92 $\pm$ 87.79 & \textbf{97.76 $\pm$ 45.68} \\
\hspace{6pt} Magnitude & \textbf{234.94 $\pm$ 33.10} & 201.71 $\pm$ 79.04 & \textbf{126.53 $\pm$ 71.54} & \textbf{78.67 $\pm$ 45.87} \\
\hspace{6pt} Magnitude Hybrid & \textbf{222.04 $\pm$ 69.39} & \textbf{173.69 $\pm$ 81.12} & 149.22 $\pm$ 53.44 & \textbf{86.86 $\pm$ 66.62} \\
\hspace{6pt} Margin & 297.03 $\pm$ 108.07 & 205.75 $\pm$ 98.25 & 146.27 $\pm$ 73.90 & \textbf{86.75 $\pm$ 85.58} \\
\hspace{6pt} VAE & \textbf{231.33 $\pm$ 40.83} & 175.57 $\pm$ 84.45 & 184.64 $\pm$ 122.74 & 117.79 $\pm$ 62.61 \\
\hspace{6pt} VAE Hybrid & 273.70 $\pm$ 88.07 & 191.05 $\pm$ 101.76 & \textbf{101.90 $\pm$ 41.17} & 99.30 $\pm$ 13.58 \\
\bottomrule
\end{tabular}}
\caption{\small Final episodic return (mean $\pm$ std) across domains and noise levels for each noise type in PEBBLE. \emph{Uniform} serves as the reference. Lower performance than uniform are shown in bold font, suggesting negative impact on performance as compared to uniform noise.}
\label{tab: episodic_return_all_noise_pebble}
\end{table*}

\begin{table*}[t]
\centering
\renewcommand{\arraystretch}{0.9}
\setlength{\tabcolsep}{3pt}
\scriptsize

\resizebox{0.78\textwidth}{!}{
\begin{tabular}{lcccc}
\toprule
\textbf{Noise Type} & \textbf{10\%} & \textbf{20\%} & \textbf{30\%} & \textbf{40\%} \\
\midrule

\multicolumn{5}{l}{\textbf{Sweep Into}} \\[1pt]
\hspace{6pt} Uniform & 849.86 $\pm$ 934.31 & 216.76 $\pm$ 206.87 & 115.22 $\pm$ 192.16 & 37.17 $\pm$ 24.27 \\
\cmidrule(lr){1-5}
\hspace{6pt} VAE & \textbf{793.19 $\pm$ 772.83} & 619.08 $\pm$ 483.38 & 307.62 $\pm$ 396.93 & 301.87 $\pm$ 285.29 \\
\hspace{6pt} VAE Hybrid & 875.10 $\pm$ 585.41 & 871.65 $\pm$ 1423.36 & 118.87 $\pm$ 185.88 & 48.88 $\pm$ 18.47 \\
\hspace{6pt} Distance & 918.72 $\pm$ 768.93 & 516.34 $\pm$ 843.76 & 382.77 $\pm$ 715.75 & 52.52 $\pm$ 23.87 \\
\hspace{6pt} Distance Hybrid & 901.72 $\pm$ 928.07 & 722.97 $\pm$ 1122.98 & \textbf{93.59 $\pm$ 59.24} & 48.79 $\pm$ 22.33 \\
\hspace{6pt} Magnitude & 1618.96 $\pm$ 1294.52 & 491.62 $\pm$ 349.10 & \textbf{68.50 $\pm$ 54.81} & 39.39 $\pm$ 37.23 \\
\hspace{6pt} Magnitude Hybrid & 1299.76 $\pm$ 1157.65 & 428.19 $\pm$ 509.99 & \textbf{102.39 $\pm$ 48.55} & 194.84 $\pm$ 413.00 \\
\hspace{6pt} Margin & 1305.12 $\pm$ 1197.61 & \textbf{186.52 $\pm$ 148.17} & \textbf{78.05 $\pm$ 61.03} & \textbf{32.24 $\pm$ 12.81} \\

\midrule
\multicolumn{5}{l}{\textbf{Hammer}} \\[1pt]
\hspace{6pt} Uniform & 2477.78 $\pm$ 462.06 & 1607.88 $\pm$ 721.93 & 437.23 $\pm$ 28.97 & 467.64 $\pm$ 24.58 \\
\cmidrule(lr){1-5}
\hspace{6pt} VAE & \textbf{2129.59 $\pm$ 297.41} & \textbf{1382.74 $\pm$ 519.15} & 591.60 $\pm$ 186.88 & 542.69 $\pm$ 110.64 \\
\hspace{6pt} VAE Hybrid & \textbf{2186.27 $\pm$ 512.39} & 1980.57 $\pm$ 658.65 & 1213.92 $\pm$ 247.62 & 762.06 $\pm$ 449.62 \\
\hspace{6pt} Distance & \textbf{2001.76 $\pm$ 630.83} & 2023.27 $\pm$ 773.97 & 598.94 $\pm$ 214.59 & 499.40 $\pm$ 107.56 \\
\hspace{6pt} Distance Hybrid & \textbf{1924.65 $\pm$ 362.52} & 2129.34 $\pm$ 702.05 & 1583.95 $\pm$ 700.43 & 640.98 $\pm$ 277.90 \\
\hspace{6pt} Magnitude & \textbf{2449.97 $\pm$ 848.02} & 1609.82 $\pm$ 537.79 & 645.99 $\pm$ 390.87 & \textbf{396.33 $\pm$ 92.83} \\
\hspace{6pt} Magnitude Hybrid & \textbf{2330.55 $\pm$ 620.98} & 1620.07 $\pm$ 1055.47 & 1135.58 $\pm$ 593.88 & \textbf{451.34 $\pm$ 28.83} \\
\hspace{6pt} Margin & \textbf{2254.86 $\pm$ 746.83} & 1811.21 $\pm$ 429.41 & 877.92 $\pm$ 308.28 & \textbf{436.89 $\pm$ 63.56} \\

\midrule
\multicolumn{5}{l}{\textbf{Button Press}} \\[1pt]
\hspace{6pt} Uniform & 2216.16 $\pm$ 1061.03 & 1103.24 $\pm$ 515.89 & 258.97 $\pm$ 142.98 & 237.30 $\pm$ 130.83 \\
\cmidrule(lr){1-5}
\hspace{6pt} VAE & 2590.83 $\pm$ 719.46 & \textbf{852.47 $\pm$ 515.68} & 452.66 $\pm$ 106.62 & 259.15 $\pm$ 112.05 \\
\hspace{6pt} VAE Hybrid & 3077.71 $\pm$ 378.37 & \textbf{1050.75 $\pm$ 318.20} & 482.41 $\pm$ 303.94 & \textbf{135.71 $\pm$ 60.48} \\
\hspace{6pt} Distance & 2608.96 $\pm$ 779.21 & 1740.86 $\pm$ 1026.32 & 743.93 $\pm$ 566.74 & \textbf{210.78 $\pm$ 144.88} \\
\hspace{6pt} Distance Hybrid & 2321.34 $\pm$ 1017.22 & 1282.39 $\pm$ 261.00 & 512.64 $\pm$ 185.74 & 276.49 $\pm$ 169.86 \\
\hspace{6pt} Magnitude & 2485.51 $\pm$ 991.42 & 1357.08 $\pm$ 711.27 & 289.80 $\pm$ 206.95 & 251.65 $\pm$ 242.40 \\
\hspace{6pt} Magnitude Hybrid & 2787.84 $\pm$ 475.11 & 1171.02 $\pm$ 536.45 & 448.80 $\pm$ 235.45 & 369.84 $\pm$ 346.82 \\
\hspace{6pt} Margin & 2445.19 $\pm$ 710.87 & 1151.25 $\pm$ 643.93 & 517.82 $\pm$ 395.00 & 293.24 $\pm$ 150.40 \\

\bottomrule
\end{tabular}}
\caption{\small Final episodic return (mean $\pm$ std) across Metaworld domains and noise levels for each noise type in PEBBLE. \emph{Uniform} serves as the reference. Lower performance than uniform are shown in bold font, suggesting negative impact on performance as compared to uniform noise.}
\label{tab:metaworld_results_pebble}
\end{table*}

%% file: chapters/tables/rune_table.tex
\begin{table*}[t]
\centering
\label{tab:rune_results}

\renewcommand{\arraystretch}{0.9}
\setlength{\tabcolsep}{3pt}
\scriptsize

\resizebox{0.78\textwidth}{!}{
\begin{tabular}{lcccc}
\toprule
\textbf{Noise Type} & \textbf{10\%} & \textbf{20\%} & \textbf{30\%} & \textbf{40\%} \\
\midrule

\multicolumn{5}{l}{\textbf{Walker Walk}} \\[1pt]
\hspace{6pt} Uniform & 868.65 $\pm$ 89.71 & 614.84 $\pm$ 256.45 & 405.52 $\pm$ 217.67 & 64.09 $\pm$ 42.74 \\
\cmidrule(lr){1-5}
\hspace{6pt} Distance & 872.19 $\pm$ 60.93 & \textbf{527.98 $\pm$ 330.82} & \textbf{266.96 $\pm$ 205.03} & 247.84 $\pm$ 128.49 \\
\hspace{6pt} Distance Hybrid & \textbf{862.25 $\pm$ 73.37} & \textbf{609.48 $\pm$ 138.46} & \textbf{359.14 $\pm$ 178.14} & 191.59 $\pm$ 183.99 \\
\hspace{6pt} Magnitude & \textbf{841.67 $\pm$ 124.90} & 698.10 $\pm$ 165.10 & \textbf{378.89 $\pm$ 204.25} & 137.16 $\pm$ 96.41 \\
\hspace{6pt} Magnitude Hybrid & 887.03 $\pm$ 67.30 & \textbf{560.82 $\pm$ 232.77} & \textbf{364.32 $\pm$ 187.09} & 160.73 $\pm$ 160.74 \\
\hspace{6pt} Margin & \textbf{797.25 $\pm$ 111.26} & \textbf{572.69 $\pm$ 194.56} & \textbf{182.38 $\pm$ 118.22} & 86.01 $\pm$ 39.85 \\
\hspace{6pt} VAE & \textbf{792.53 $\pm$ 186.72} & \textbf{573.67 $\pm$ 259.68} & 447.46 $\pm$ 41.43 & 214.27 $\pm$ 276.59 \\
\hspace{6pt} VAE Hybrid & \textbf{857.09 $\pm$ 121.74} & \textbf{601.84 $\pm$ 228.70} & \textbf{210.91 $\pm$ 145.39} & 114.17 $\pm$ 68.85 \\
\midrule
\multicolumn{5}{l}{\textbf{HalfCheetah Run}} \\[1pt]
\hspace{6pt} Uniform & 655.07 $\pm$ 64.22 & 632.45 $\pm$ 40.29 & 455.83 $\pm$ 157.67 & 295.77 $\pm$ 70.19 \\
\cmidrule(lr){1-5}
\hspace{6pt} Distance & 666.37 $\pm$ 72.98 & \textbf{522.91 $\pm$ 157.42} & \textbf{437.13 $\pm$ 212.42} & \textbf{206.39 $\pm$ 136.90} \\
\hspace{6pt} Distance Hybrid & \textbf{624.90 $\pm$ 90.56} & \textbf{600.47 $\pm$ 70.61} & \textbf{352.80 $\pm$ 223.45} & \textbf{240.21 $\pm$ 168.94} \\
\hspace{6pt} Magnitude & 673.10 $\pm$ 85.23 & \textbf{552.78 $\pm$ 106.28} & \textbf{370.04 $\pm$ 159.63} & \textbf{122.74 $\pm$ 106.93} \\
\hspace{6pt} Magnitude Hybrid & \textbf{605.60 $\pm$ 84.58} & \textbf{520.20 $\pm$ 88.50} & \textbf{413.27 $\pm$ 217.14} & \textbf{118.95 $\pm$ 80.13} \\
\hspace{6pt} Margin & \textbf{654.13 $\pm$ 82.59} & \textbf{612.84 $\pm$ 140.67} & \textbf{413.22 $\pm$ 39.75} & \textbf{210.95 $\pm$ 109.38} \\
\hspace{6pt} VAE & \textbf{655.03 $\pm$ 59.71} & \textbf{592.66 $\pm$ 113.31} & 566.49 $\pm$ 147.05 & 323.44 $\pm$ 141.84 \\
\hspace{6pt} VAE Hybrid & \textbf{620.35 $\pm$ 152.67} & \textbf{550.14 $\pm$ 234.27} & \textbf{404.10 $\pm$ 76.72} & \textbf{155.54 $\pm$ 108.37} \\
\midrule
\multicolumn{5}{l}{\textbf{Quadruped Walk}} \\[1pt]
\hspace{6pt} Uniform & 246.88 $\pm$ 59.14 & 191.66 $\pm$ 54.63 & 139.23 $\pm$ 100.69 & 112.65 $\pm$ 18.42 \\
\cmidrule(lr){1-5}
\hspace{6pt} Distance & 350.40 $\pm$ 138.95 & 221.56 $\pm$ 65.92 & \textbf{98.53 $\pm$ 62.05} & \textbf{76.47 $\pm$ 54.44} \\
\hspace{6pt} Distance Hybrid & 301.27 $\pm$ 216.56 & 230.85 $\pm$ 152.66 & 141.23 $\pm$ 72.54 & \textbf{57.70 $\pm$ 20.52} \\
\hspace{6pt} Magnitude & \textbf{199.87 $\pm$ 29.80} & \textbf{182.91 $\pm$ 29.52} & \textbf{115.77 $\pm$ 57.40} & \textbf{61.38 $\pm$ 59.06} \\
\hspace{6pt} Magnitude Hybrid & 323.99 $\pm$ 111.68 & \textbf{137.25 $\pm$ 29.78} & \textbf{119.37 $\pm$ 37.63} & \textbf{83.85 $\pm$ 67.73} \\
\hspace{6pt} Margin & \textbf{226.09 $\pm$ 64.05} & \textbf{174.89 $\pm$ 47.25} & \textbf{109.15 $\pm$ 56.93} & \textbf{82.75 $\pm$ 31.57} \\
\hspace{6pt} VAE & 290.12 $\pm$ 188.06 & \textbf{164.01 $\pm$ 93.32} & 142.26 $\pm$ 45.44 & \textbf{78.03 $\pm$ 36.12} \\
\hspace{6pt} VAE Hybrid & \textbf{213.55 $\pm$ 31.39} & 217.90 $\pm$ 49.01 & \textbf{114.31 $\pm$ 63.71} & \textbf{101.51 $\pm$ 39.69} \\
\bottomrule
\end{tabular}}
\caption{\small Final episodic return (mean $\pm$ std) across domains and noise levels for each noise type in RUNE. \emph{Uniform} serves as the reference. Lower performance are shown in bold font, suggesting negative impact on performance for the specific noise types.}
\label{tab: episodic_return_all_noise_rune}
\end{table*}

\begin{table*}[t]
\centering
\renewcommand{\arraystretch}{0.9}
\setlength{\tabcolsep}{3pt}
\scriptsize

\resizebox{0.78\textwidth}{!}{
\begin{tabular}{lcccc}
\toprule
\textbf{Noise Type} & \textbf{10\%} & \textbf{20\%} & \textbf{30\%} & \textbf{40\%} \\
\midrule

\multicolumn{5}{l}{\textbf{Sweep Into}} \\[1pt]
\hspace{6pt} Uniform & 667.25 $\pm$ 564.47 & 250.31 $\pm$ 269.05 & 108.19 $\pm$ 129.04 & 39.84 $\pm$ 14.96 \\
\cmidrule(lr){1-5}
\hspace{6pt} VAE & \textbf{471.58 $\pm$ 313.49} & 418.79 $\pm$ 726.23 & \textbf{78.23 $\pm$ 45.76} & 50.91 $\pm$ 25.73 \\
\hspace{6pt} VAE Hybrid & 1720.87 $\pm$ 1225.03 & \textbf{136.14 $\pm$ 95.23} & 121.76 $\pm$ 131.23 & 78.08 $\pm$ 56.53 \\
\hspace{6pt} Distance & 958.13 $\pm$ 859.23 & 434.54 $\pm$ 528.39 & 211.67 $\pm$ 358.65 & 49.01 $\pm$ 35.48 \\
\hspace{6pt} Distance Hybrid & 1325.24 $\pm$ 1137.57 & \textbf{197.32 $\pm$ 161.30} & 590.26 $\pm$ 893.29 & 51.93 $\pm$ 11.15 \\
\hspace{6pt} Magnitude & 757.31 $\pm$ 757.51 & 271.68 $\pm$ 196.58 & 342.38 $\pm$ 747.10 & 41.36 $\pm$ 28.24 \\
\hspace{6pt} Magnitude Hybrid & \textbf{612.21 $\pm$ 505.92} & 991.59 $\pm$ 1166.86 & \textbf{53.01 $\pm$ 14.05} & 75.43 $\pm$ 87.30 \\
\hspace{6pt} Margin & 761.92 $\pm$ 549.88 & \textbf{216.67 $\pm$ 224.48} & \textbf{53.82 $\pm$ 21.02} & 47.21 $\pm$ 27.93 \\

\midrule
\multicolumn{5}{l}{\textbf{Hammer}} \\[1pt]
\hspace{6pt} Uniform & 2179.51 $\pm$ 571.59 & 1318.92 $\pm$ 472.35 & 441.17 $\pm$ 53.10 & 416.66 $\pm$ 45.20 \\
\cmidrule(lr){1-5}
\hspace{6pt} VAE & \textbf{2123.99 $\pm$ 788.17} & 1704.72 $\pm$ 540.86 & 640.42 $\pm$ 223.18 & 483.19 $\pm$ 22.97 \\
\hspace{6pt} VAE Hybrid & \textbf{1785.45 $\pm$ 62.12} & 1479.93 $\pm$ 651.85 & 891.98 $\pm$ 377.08 & 512.44 $\pm$ 103.60 \\
\hspace{6pt} Distance & 2468.02 $\pm$ 596.89 & \textbf{1308.66 $\pm$ 582.98} & 964.97 $\pm$ 418.01 & 827.16 $\pm$ 498.82 \\
\hspace{6pt} Distance Hybrid & \textbf{1953.57 $\pm$ 316.10} & 1448.22 $\pm$ 677.94 & 858.42 $\pm$ 409.29 & 572.31 $\pm$ 210.78 \\
\hspace{6pt} Magnitude & 2368.03 $\pm$ 1072.31 & 1648.12 $\pm$ 745.35 & 1208.85 $\pm$ 682.54 & 424.20 $\pm$ 92.08 \\
\hspace{6pt} Magnitude Hybrid & \textbf{1590.64 $\pm$ 753.73} & 1651.40 $\pm$ 814.99 & 1114.22 $\pm$ 720.79 & 440.46 $\pm$ 37.90 \\
\hspace{6pt} Margin & 2220.12 $\pm$ 754.99 & 1603.75 $\pm$ 316.39 & 814.95 $\pm$ 387.39 & 499.98 $\pm$ 232.72 \\

\midrule
\multicolumn{5}{l}{\textbf{Button Press}} \\[1pt]
\hspace{6pt} Uniform & 2569.26 $\pm$ 795.67 & 565.37 $\pm$ 364.69 & 234.48 $\pm$ 205.11 & 184.31 $\pm$ 152.93 \\
\cmidrule(lr){1-5}
\hspace{6pt} VAE & 2839.56 $\pm$ 204.93 & 1072.83 $\pm$ 822.61 & 603.80 $\pm$ 225.08 & \textbf{169.83 $\pm$ 121.91} \\
\hspace{6pt} VAE Hybrid & \textbf{2269.25 $\pm$ 791.08} & 1314.33 $\pm$ 410.18 & 674.30 $\pm$ 273.01 & 392.98 $\pm$ 179.70 \\
\hspace{6pt} Distance & 2693.37 $\pm$ 629.93 & 1498.28 $\pm$ 675.44 & 591.09 $\pm$ 345.31 & 250.25 $\pm$ 173.06 \\
\hspace{6pt} Distance Hybrid & 2865.59 $\pm$ 350.40 & 1382.32 $\pm$ 591.00 & 464.31 $\pm$ 345.31 & 300.81 $\pm$ 161.28 \\
\hspace{6pt} Magnitude & \textbf{2257.21 $\pm$ 1033.99} & 755.17 $\pm$ 671.10 & 450.67 $\pm$ 269.75 & 246.13 $\pm$ 201.87 \\
\hspace{6pt} Magnitude Hybrid & 3029.87 $\pm$ 185.95 & 1029.89 $\pm$ 968.92 & 476.30 $\pm$ 379.50 & 221.45 $\pm$ 217.35 \\
\hspace{6pt} Margin & 3169.02 $\pm$ 369.92 & 1072.48 $\pm$ 922.82 & 377.57 $\pm$ 302.20 & 258.88 $\pm$ 173.49 \\

\bottomrule
\end{tabular}}
\caption{\small Final episodic return (mean $\pm$ std) across Metaworld domains and noise levels for each noise type in RUNE. \emph{Uniform} serves as the reference. Lower performance are shown in bold font, suggesting negative impact on performance for the specific noise types.}
\label{tab:metaworld_results_rune}
\end{table*}

%% file: chapters/tables/surf_table.tex
\begin{table*}[t]
\centering
\label{tab:surf_results}

\renewcommand{\arraystretch}{0.9}
\setlength{\tabcolsep}{3pt}
\scriptsize

\resizebox{0.78\textwidth}{!}{
\begin{tabular}{lcccc}
\toprule
\textbf{Noise Type} & \textbf{10\%} & \textbf{20\%} & \textbf{30\%} & \textbf{40\%} \\
\midrule

\multicolumn{5}{l}{\textbf{Walker Walk}} \\[1pt]
\hspace{6pt} Uniform & 716.37 $\pm$ 64.08 & 423.93 $\pm$ 216.66 & 93.23 $\pm$ 42.56 & 46.16 $\pm$ 22.27 \\
\cmidrule(lr){1-5}
\hspace{6pt} Distance & \textbf{512.30 $\pm$ 305.88} & \textbf{155.27 $\pm$ 54.47} & 132.81 $\pm$ 107.35 & 57.92 $\pm$ 28.27 \\
\hspace{6pt} Distance Hybrid & \textbf{565.10 $\pm$ 240.22} & \textbf{250.31 $\pm$ 133.96} & 175.42 $\pm$ 153.50 & 49.65 $\pm$ 29.72 \\
\hspace{6pt} Magnitude & \textbf{684.79 $\pm$ 90.09} & \textbf{326.01 $\pm$ 275.94} & 228.13 $\pm$ 223.93 & \textbf{42.65 $\pm$ 29.01} \\
\hspace{6pt} Magnitude Hybrid & 744.81 $\pm$ 26.38 & 505.69 $\pm$ 182.98 & 107.94 $\pm$ 64.15 & 53.97 $\pm$ 25.81 \\
\hspace{6pt} Margin & \textbf{568.09 $\pm$ 230.35} & 452.37 $\pm$ 240.03 & 194.67 $\pm$ 151.56 & 57.46 $\pm$ 23.89 \\
\hspace{6pt} VAE & \textbf{518.60 $\pm$ 166.12} & \textbf{191.00 $\pm$ 91.88} & \textbf{53.22 $\pm$ 25.20} & 180.25 $\pm$ 234.04 \\
\hspace{6pt} VAE Hybrid & \textbf{281.23 $\pm$ 98.34} & \textbf{294.54 $\pm$ 194.18} & \textbf{72.41 $\pm$ 20.44} & \textbf{21.46 $\pm$ 6.78} \\
\midrule
\multicolumn{5}{l}{\textbf{HalfCheetah Run}} \\[1pt]
\hspace{6pt} Uniform & 395.83 $\pm$ 100.52 & 339.57 $\pm$ 104.88 & 200.69 $\pm$ 68.49 & 70.58 $\pm$ 87.11 \\
\cmidrule(lr){1-5}
\hspace{6pt} Distance & 416.95 $\pm$ 21.60 & \textbf{298.75 $\pm$ 74.44} & \textbf{157.93 $\pm$ 152.67} & 93.46 $\pm$ 99.32 \\
\hspace{6pt} Distance Hybrid & 422.66 $\pm$ 72.90 & \textbf{290.50 $\pm$ 134.04} & \textbf{195.14 $\pm$ 134.44} & 73.17 $\pm$ 94.95 \\
\hspace{6pt} Magnitude & \textbf{354.91 $\pm$ 147.86} & \textbf{295.24 $\pm$ 124.22} & 206.34 $\pm$ 95.00 & 135.76 $\pm$ 131.29 \\
\hspace{6pt} Magnitude Hybrid & \textbf{333.61 $\pm$ 57.12} & \textbf{302.97 $\pm$ 99.18} & \textbf{155.84 $\pm$ 179.45} & 95.91 $\pm$ 99.87 \\
\hspace{6pt} Margin & \textbf{364.21 $\pm$ 114.64} & \textbf{261.69 $\pm$ 123.18} & 213.03 $\pm$ 127.33 & 170.67 $\pm$ 101.16 \\
\hspace{6pt} VAE & \textbf{389.22 $\pm$ 94.80} & \textbf{299.94 $\pm$ 215.71} & \textbf{195.10 $\pm$ 139.66} & 75.47 $\pm$ 71.62 \\
\hspace{6pt} VAE Hybrid & 397.05 $\pm$ 70.36 & \textbf{259.61 $\pm$ 180.34} & \textbf{160.41 $\pm$ 70.65} & 119.20 $\pm$ 88.47 \\
\midrule
\multicolumn{5}{l}{\textbf{Quadruped Walk}} \\[1pt]
\hspace{6pt} Uniform & 200.57 $\pm$ 50.29 & 145.11 $\pm$ 52.48 & 104.77 $\pm$ 46.51 & 66.51 $\pm$ 67.70 \\
\cmidrule(lr){1-5}
\hspace{6pt} Distance & 272.89 $\pm$ 131.87 & 203.41 $\pm$ 59.15 & 173.03 $\pm$ 56.99 & 105.48 $\pm$ 30.09 \\
\hspace{6pt} Distance Hybrid & 240.09 $\pm$ 85.41 & 266.30 $\pm$ 175.44 & 227.87 $\pm$ 166.72 & 111.21 $\pm$ 38.80 \\
\hspace{6pt} Magnitude & \textbf{178.00 $\pm$ 21.09} & 211.43 $\pm$ 16.60 & 113.60 $\pm$ 73.15 & \textbf{49.37 $\pm$ 42.09} \\
\hspace{6pt} Magnitude Hybrid & 254.28 $\pm$ 113.33 & 218.82 $\pm$ 67.42 & 149.74 $\pm$ 26.18 & 127.48 $\pm$ 49.43 \\
\hspace{6pt} Margin & 226.09 $\pm$ 58.65 & \textbf{127.41 $\pm$ 33.74} & 148.31 $\pm$ 28.05 & 132.57 $\pm$ 45.04 \\
\hspace{6pt} VAE & \textbf{183.51 $\pm$ 76.46} & 196.50 $\pm$ 26.72 & 126.78 $\pm$ 80.61 & 120.07 $\pm$ 64.52 \\
\hspace{6pt} VAE Hybrid & 205.63 $\pm$ 15.96 & 168.28 $\pm$ 54.96 & 164.09 $\pm$ 96.07 & 85.13 $\pm$ 48.19 \\
\bottomrule
\end{tabular}}
\caption{\small Final episodic return (mean $\pm$ std) across domains and noise levels for each noise type in SURF. \emph{Uniform} serves as the reference. Lower performance are shown in bold font, suggesting negative impact on performance for the specific noise types.}
\label{tab: episodic_return_all_noise_surf}
\end{table*}

\begin{table*}[t]
\centering
\renewcommand{\arraystretch}{0.9}
\setlength{\tabcolsep}{3pt}
\scriptsize

\resizebox{0.78\textwidth}{!}{
\begin{tabular}{lcccc}
\toprule
\textbf{Noise Type} & \textbf{10\%} & \textbf{20\%} & \textbf{30\%} & \textbf{40\%} \\
\midrule

\multicolumn{5}{l}{\textbf{Sweep Into}} \\[1pt]
\hspace{6pt} Uniform & 1431.99 $\pm$ 1452.20 & 493.41 $\pm$ 512.23 & 178.83 $\pm$ 204.26 & 26.89 $\pm$ 16.11 \\
\cmidrule(lr){1-5}
\hspace{6pt} VAE & \textbf{1137.19 $\pm$ 1041.72} & 702.86 $\pm$ 792.01 & 502.87 $\pm$ 509.21 & 417.00 $\pm$ 803.13 \\
\hspace{6pt} VAE Hybrid & 1682.28 $\pm$ 1518.68 & \textbf{328.61 $\pm$ 331.62} & 384.68 $\pm$ 713.78 & 52.80 $\pm$ 36.30 \\
\hspace{6pt} Distance & \textbf{714.91 $\pm$ 383.45} & \textbf{351.72 $\pm$ 336.27} & 241.79 $\pm$ 242.24 & 224.78 $\pm$ 363.34 \\
\hspace{6pt} Distance Hybrid & \textbf{896.52 $\pm$ 879.64} & \textbf{456.31 $\pm$ 502.16} & 252.60 $\pm$ 197.73 & 130.04 $\pm$ 189.40 \\
\hspace{6pt} Magnitude & \textbf{1063.00 $\pm$ 1060.45} & \textbf{382.55 $\pm$ 418.72} & \textbf{70.11 $\pm$ 45.61} & 45.23 $\pm$ 35.27 \\
\hspace{6pt} Magnitude Hybrid & \textbf{436.77 $\pm$ 561.33} & \textbf{492.22 $\pm$ 292.66} & \textbf{54.50 $\pm$ 8.52} & 128.93 $\pm$ 235.86 \\
\hspace{6pt} Margin & \textbf{771.91 $\pm$ 640.50} & \textbf{325.88 $\pm$ 235.29} & \textbf{65.77 $\pm$ 20.64} & 224.89 $\pm$ 462.87 \\

\midrule
\multicolumn{5}{l}{\textbf{Hammer}} \\[1pt]
\hspace{6pt} Uniform & 2080.22 $\pm$ 664.79 & 1631.21 $\pm$ 1026.09 & 1015.85 $\pm$ 224.94 & 446.96 $\pm$ 53.81 \\
\cmidrule(lr){1-5}
\hspace{6pt} VAE & 2131.04 $\pm$ 607.77 & 1878.26 $\pm$ 749.53 & 1189.68 $\pm$ 413.89 & 611.58 $\pm$ 304.02 \\
\hspace{6pt} VAE Hybrid & \textbf{1177.91 $\pm$ 557.98} & 1718.92 $\pm$ 754.42 & 1050.49 $\pm$ 586.76 & 585.33 $\pm$ 154.28 \\
\hspace{6pt} Distance & 2245.61 $\pm$ 468.73 & \textbf{1581.21 $\pm$ 629.85} & 1075.06 $\pm$ 367.66 & 914.88 $\pm$ 671.30 \\
\hspace{6pt} Distance Hybrid & \textbf{1743.08 $\pm$ 497.99} & 1904.66 $\pm$ 645.05 & 1146.99 $\pm$ 322.25 & 522.00 $\pm$ 85.30 \\
\hspace{6pt} Magnitude & 2427.40 $\pm$ 530.34 & 1912.80 $\pm$ 752.47 & 1233.04 $\pm$ 712.92 & 509.10 $\pm$ 111.07 \\
\hspace{6pt} Magnitude Hybrid & \textbf{1734.29 $\pm$ 774.34} & 1929.13 $\pm$ 869.37 & 1512.58 $\pm$ 327.30 & 472.07 $\pm$ 110.64 \\
\hspace{6pt} Margin & \textbf{1938.88 $\pm$ 585.17} & 1822.26 $\pm$ 844.88 & 1223.05 $\pm$ 729.58 & 467.47 $\pm$ 66.09 \\

\midrule
\multicolumn{5}{l}{\textbf{Button Press}} \\[1pt]
\hspace{6pt} Uniform & 2790.45 $\pm$ 511.27 & 1441.60 $\pm$ 555.04 & 708.87 $\pm$ 450.76 & 155.69 $\pm$ 76.94 \\
\cmidrule(lr){1-5}
\hspace{6pt} VAE & \textbf{2744.76 $\pm$ 497.44} & \textbf{1318.73 $\pm$ 478.23} & \textbf{674.97 $\pm$ 238.12} & 316.44 $\pm$ 140.32 \\
\hspace{6pt} VAE Hybrid & \textbf{2678.26 $\pm$ 860.52} & \textbf{1171.44 $\pm$ 291.27} & 822.72 $\pm$ 336.57 & 345.95 $\pm$ 159.48 \\
\hspace{6pt} Distance & \textbf{2720.18 $\pm$ 579.48} & \textbf{1327.36 $\pm$ 524.26} & 1004.11 $\pm$ 730.09 & 549.54 $\pm$ 292.19 \\
\hspace{6pt} Distance Hybrid & 2935.35 $\pm$ 737.64 & \textbf{1148.85 $\pm$ 192.01} & 1027.23 $\pm$ 492.53 & 280.48 $\pm$ 66.42 \\
\hspace{6pt} Magnitude & 2916.96 $\pm$ 412.77 & 1748.60 $\pm$ 810.14 & \textbf{338.14 $\pm$ 186.59} & 275.38 $\pm$ 145.02 \\
\hspace{6pt} Magnitude Hybrid & \textbf{2774.09 $\pm$ 416.62} & 1737.51 $\pm$ 527.32 & 788.79 $\pm$ 166.86 & \textbf{138.04 $\pm$ 93.28} \\
\hspace{6pt} Margin & \textbf{2705.49 $\pm$ 537.65} & 1530.72 $\pm$ 672.95 & \textbf{638.84 $\pm$ 159.96} & 345.13 $\pm$ 210.69 \\

\bottomrule
\end{tabular}}
\caption{\small Final episodic return (mean $\pm$ std) across Metaworld domains and noise levels for each noise type in SURF. \emph{Uniform} serves as the reference. Lower performance are shown in bold font, suggesting negative impact on performance for the specific noise types.}
\label{tab:metaworld_results_surf}
\end{table*}

%% file: chapters/appendix_downweight.tex
\section{Mitigating the FDN noise in PbRL}

\begin{figure*}[!t]
    \centering
    \includegraphics[width=0.24\linewidth]{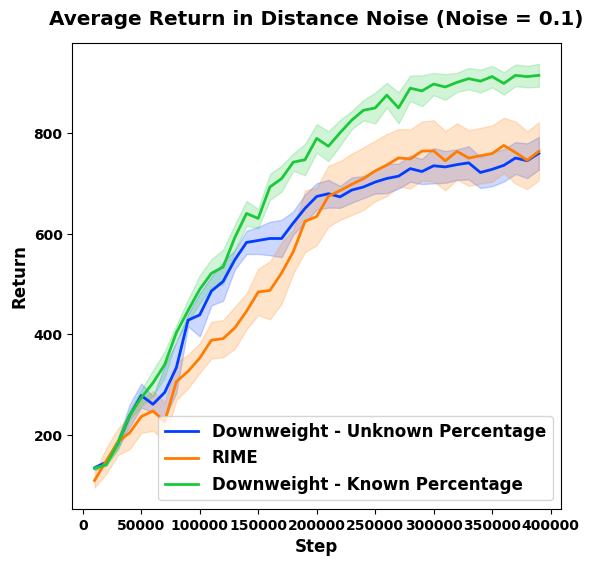}
    \includegraphics[width=0.24\linewidth]{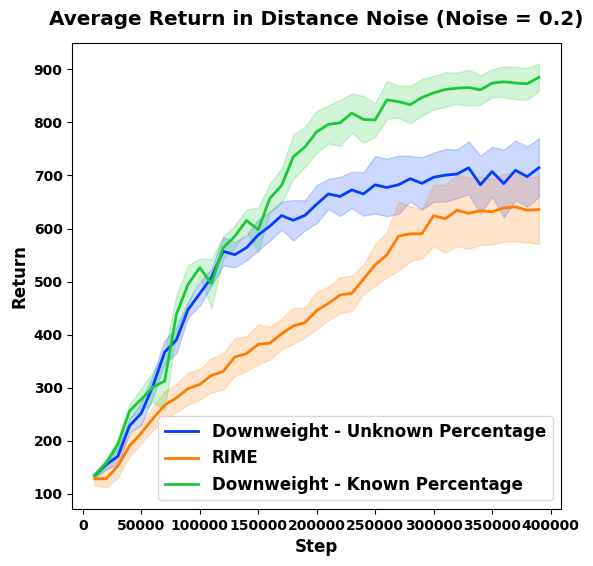}
    \includegraphics[width=0.24\linewidth]{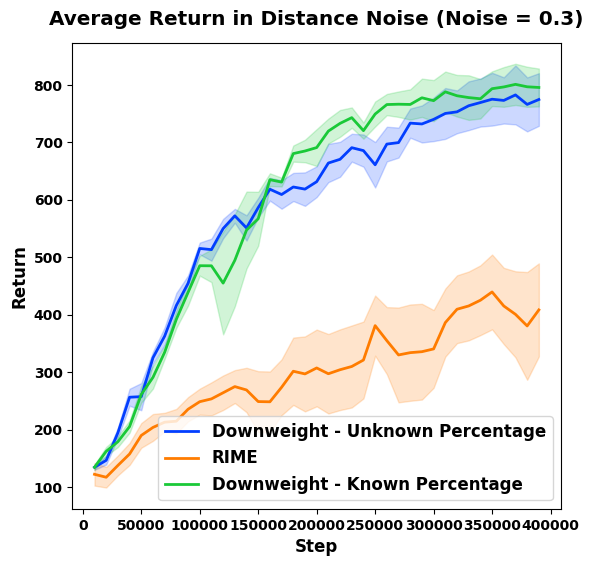}
    \includegraphics[width=0.24\linewidth]{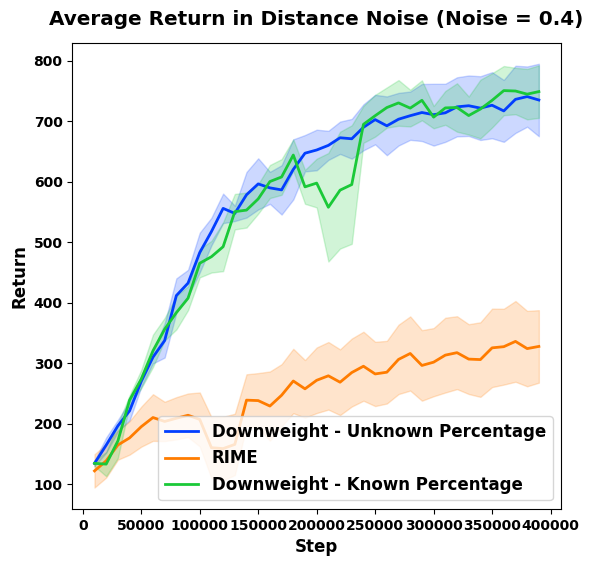}
    \caption{Trajectory similarity Noise with down-weighting mechanism across different noise scale, in Walker Walk.}
    \label{fig:percentile_walker}
\end{figure*}

As we found in previous experiments, state-of-the-art noise-robust PbRL algorithms like RIME suffer a performance downgrade in FDNs, with a drastically different pattern with uniform noise. 
We conducted extra experiments on a simple yet effective strategy to combat FDN by simply down-weighting the losses of preference items where FDNs are more likely to happen on top of RIME. In other words, our assumption here is that we know what kind of FDN the teacher provides which is rather a strict assumption. For example, if we know we have trajectory similarity noise, we can downweight or even completely remove losses sourced from preferences over similar trajectories.
The results are shown in Figure~\ref{fig:percentile_walker}. "Unknown Noise Percentage" indicates the case where we do not know the proportion of the noise percentage and we only learn from 50\%  of preferences over trajectory pairs that's least likely to have noise, which is a extremely conservative strategy. "Known Noise Percentage", on the other hand is where we known the ground truth noise percentage and we learn from (1 - noise percentage)\% of probable clean preferences.
We can find that "Known Noise Percentage" constantly outperform the baseline RIME, warranting the effectiveness of this downweighting mechanism. 
Furthermore, it is observed that being conservative and only learn from top 50\% clean candidates is beneficial in high noise cases like 30\% and 40\% noise. 
confirms that RIME is not able properly filter the noisy preference and learning from these low-quality data is harming the learning performance more than completely ignoring them.
Here to conclude, while FDN shows a completely different noise pattern, it is possible to mitigate with FDN with our down-weighting mechanism within known percentage of noise and sometime even proportion of noise. We leave further exploration of FDN specific algorithms as future work.

%% file: main.bib
@InProceedings{wang2024,
          title = 	 {RL-VLM-F: Reinforcement Learning from Vision Language Foundation Model Feedback},
          author =       {Wang, Yufei and Sun, Zhanyi and Zhang, Jesse and Xian, Zhou and Biyik, Erdem and Held, David and Erickson, Zackory},
          booktitle = 	 {Proceedings of the 41th International Conference on Machine Learning},
          year = 	 {2024}
        }

@article{christiano2017deep,
  title={Deep reinforcement learning from human preferences},
  author={Christiano, Paul F and Leike, Jan and Brown, Tom and Martic, Miljan and Legg, Shane and Amodei, Dario},
  journal={Advances in neural information processing systems},
  volume={30},
  year={2017}
}

@article{Cheng2024RIME,
  title={RIME: Robust Preference-based Reinforcement Learning with Noisy Preferences},
  author={Jie Cheng and Gang Xiong and Xingyuan Dai and Qinghai Miao and Yisheng Lv and Fei-Yue Wang},
  journal={arXiv preprint arXiv:2402.17257},
  year={2024},
  url={https://arxiv.org/abs/2402.17257}
}

@article{lee2021pebble,
  title={Pebble: Feedback-efficient interactive reinforcement learning via relabeling experience and unsupervised pre-training},
  author={Lee, Kimin and Smith, Laura and Abbeel, Pieter},
  journal={arXiv preprint arXiv:2106.05091},
  year={2021}
}

@article{ouyang2022training,
  title={Training language models to follow instructions with human feedback},
  author={Ouyang, Long and Wu, Jeffrey and Jiang, Xu and Almeida, Diogo and Wainwright, Carroll and Mishkin, Pamela and Zhang, Chong and Agarwal, Sandhini and Slama, Katarina and Ray, Alex and others},
  journal={Advances in neural information processing systems},
  volume={35},
  pages={27730--27744},
  year={2022}
}

@article{rafailov2023direct,
  title={Direct preference optimization: Your language model is secretly a reward model},
  author={Rafailov, Rafael and Sharma, Archit and Mitchell, Eric and Manning, Christopher D and Ermon, Stefano and Finn, Chelsea},
  journal={Advances in neural information processing systems},
  volume={36},
  pages={53728--53741},
  year={2023}
}

@article{tassa2018dmcontrol,
  title={DeepMind Control Suite},
  author={Tassa, Yuval and Doron, Yotam and Muldal, Alistair and Erez, Tom and Li, Yazhe and de Las Casas, Diego and Budden, David and Abdolmaleki, Abbas and Merel, Josh and Lefrancq, Andrew and Lillicrap, Timothy and Riedmiller, Martin},
  journal={arXiv preprint arXiv:1801.00690},
  year={2018},
  url={https://arxiv.org/abs/1801.00690}
}

@article{li2025well,
  title={How well can LLMs provide planning feedback in grounded environments?},
  author={Li, Yuxuan and Zhong, Victor},
  journal={arXiv preprint arXiv:2509.09790},
  year={2025}
}

@article{wang2024rl,
  title={Rl-vlm-f: Reinforcement learning from vision language foundation model feedback},
  author={Wang, Yufei and Sun, Zhanyi and Zhang, Jesse and Xian, Zhou and Biyik, Erdem and Held, David and Erickson, Zackory},
  journal={arXiv preprint arXiv:2402.03681},
  year={2024}
}

@article{tu2024online,
  title={Online Preference-based Reinforcement Learning with Self-augmented Feedback from Large Language Model},
  author={Tu, Songjun and Sun, Jingbo and Zhang, Qichao and Lan, Xiangyuan and Zhao, Dongbin},
  journal={The 24th International Conference on Autonomous Agents and Multi-Agent Systems, AAMAS-2025},
  year={2024}
}

@article{lee2021bpref,
  title        = {B-Pref: Benchmarking Preference-Based Reinforcement Learning},
  author       = {Kimin Lee and Laura Smith and Anca Dragan and Pieter Abbeel},
  journal      = {CoRR},
  volume       = {abs/2111.03026},
  year         = {2021},
  url          = {https://arxiv.org/abs/2111.03026}
}

@inproceedings{huang2025trend,
  title={TREND: Tri-teaching for Robust Preference-based Reinforcement Learning with Demonstrations},
  author={Huang, Shuaiyi and Levy, Mara and Gupta, Anubhav and Ekpo, Daniel and Zheng, Ruijie and Shrivastava, Abhinav},
  booktitle={Proceedings of the IEEE International Conference on Robotics and Automation (ICRA)},
  year={2025}
}

@article{mirhoseini2021graph,
  title={A graph placement methodology for fast chip design},
  author={Mirhoseini, Azalia and Goldie, Anna and Yazgan, Mustafa and Jiang, Joe Wenjie and Songhori, Ebrahim and Wang, Shen and Lee, Young-Joon and Johnson, Eric and Pathak, Omkar and Nova, Azade and others},
  journal={Nature},
  volume={594},
  number={7862},
  pages={207--212},
  year={2021},
  publisher={Nature Publishing Group UK London}
}

@article{janjua2024gvfs,
  title={GVFs in the real world: making predictions online for water treatment},
  author={Janjua, Muhammad Kamran and Shah, Haseeb and White, Martha and Miahi, Erfan and Machado, Marlos C and White, Adam},
  journal={Machine Learning},
  volume={113},
  number={8},
  pages={5151--5181},
  year={2024},
  publisher={Springer}
}

@article{wurman2022outracing,
  title={Outracing champion Gran Turismo drivers with deep reinforcement learning},
  author={Wurman, Peter R and Barrett, Samuel and Kawamoto, Kenta and MacGlashan, James and Subramanian, Kaushik and Walsh, Thomas J and Capobianco, Roberto and Devlic, Alisa and Eckert, Franziska and Fuchs, Florian and others},
  journal={Nature},
  volume={602},
  number={7896},
  pages={223--228},
  year={2022},
  publisher={Nature Publishing Group UK London}
}

@article{lakhan2023drlbts,
  title={DRLBTS: Deep reinforcement learning-aware blockchain-based healthcare system},
  author={Lakhan, Abdullah and Mohammed, Mazin Abed and Nedoma, Jan and Martinek, Radek and Tiwari, Prayag and Kumar, Neeraj},
  journal={Scientific Reports},
  volume={13},
  number={1},
  pages={4124},
  year={2023},
  publisher={Nature Publishing Group UK London}
}

@inproceedings{booth2023perils,
  title        = {The perils of trial-and-error reward design: misdesign through overfitting and invalid task specifications},
  author       = {Booth, Serena and Knox, W. Bradley and Shah, Julie and Niekum, Scott and Stone, Peter and Allievi, Alessandro},
  booktitle    = {Proceedings of the AAAI Conference on Artificial Intelligence},
  volume       = {37},
  pages        = {5920--5929},
  year         = {2023}
}

@article{amodei2016concrete,
  title={Concrete problems in AI safety},
  author={Amodei, Dario and Olah, Chris and Steinhardt, Jacob and Christiano, Paul and Schulman, John and Man{\'e}, Dan},
  journal={arXiv preprint arXiv:1606.06565},
  year={2016}
}

@article{pan2022effects,
  title={The effects of reward misspecification: Mapping and mitigating misaligned models},
  author={Pan, Alexander and Bhatia, Kush and Steinhardt, Jacob},
  journal={arXiv preprint arXiv:2201.03544},
  year={2022}
}

@article{ibarz2021train,
  title={How to train your robot with deep reinforcement learning: lessons we have learned},
  author={Ibarz, Julian and Tan, Jie and Finn, Chelsea and Kalakrishnan, Mrinal and Pastor, Peter and Levine, Sergey},
  journal={The International Journal of Robotics Research},
  volume={40},
  number={4-5},
  pages={698--721},
  year={2021},
  publisher={SAGE Publications Sage UK: London, England}
}

@article{knox2023reward,
  title={Reward (mis) design for autonomous driving},
  author={Knox, W Bradley and Allievi, Alessandro and Banzhaf, Holger and Schmitt, Felix and Stone, Peter},
  journal={Artificial Intelligence},
  volume={316},
  pages={103829},
  year={2023},
  publisher={Elsevier}
}

@inproceedings{DBLP:conf/iclr/ParkSSLAL22,
  author       = {Jongjin Park and
                  Younggyo Seo and
                  Jinwoo Shin and
                  Honglak Lee and
                  Pieter Abbeel and
                  Kimin Lee},
  title        = {{SURF:} Semi-supervised Reward Learning with Data Augmentation for
                  Feedback-efficient Preference-based Reinforcement Learning},
  booktitle    = {The Tenth International Conference on Learning Representations, {ICLR}
                  2022, Virtual Event, April 25-29, 2022},
  publisher    = {OpenReview.net},
  year         = {2022},
  url          = {https://openreview.net/forum?id=TfhfZLQ2EJO},
  bibsource    = {dblp computer science bibliography, https://dblp.org}
}

@inproceedings{feng2025duo,
  title        = {DUO: Diverse, Uncertain, On-Policy Query Generation and Selection for Reinforcement Learning from Human Feedback},
  author       = {Feng, Xuening and Jiang, Zhaohui and Kaufmann, Timo and Xu, Puchen and H{\"u}llermeier, Eyke and Weng, Paul and Zhu, Yifei},
  booktitle    = {Proceedings of the AAAI Conference on Artificial Intelligence},
  volume       = {39},
  pages        = {16604--16612},
  year         = {2025}
}

@inproceedings{DBLP:conf/iclr/LiangSLA22,
  author       = {Xinran Liang and
                  Katherine Shu and
                  Kimin Lee and
                  Pieter Abbeel},
  title        = {Reward Uncertainty for Exploration in Preference-based Reinforcement
                  Learning},
  booktitle    = {The Tenth International Conference on Learning Representations, {ICLR}
                  2022, Virtual Event, April 25-29, 2022},
  publisher    = {OpenReview.net},
  year         = {2022},
  url          = {https://openreview.net/forum?id=OWZVD-l-ZrC},
  bibsource    = {dblp computer science bibliography, https://dblp.org}
}

@inproceedings{DBLP:conf/ijcai/Xue0YX24,
  author       = {Wanqi Xue and
                  Bo An and
                  Shuicheng Yan and
                  Zhongwen Xu},
  title        = {Reinforcement Learning from Diverse Human Preferences},
  booktitle    = {Proceedings of the Thirty-Third International Joint Conference on
                  Artificial Intelligence, {IJCAI} 2024, Jeju, South Korea, August 3-9,
                  2024},
  pages        = {5298--5306},
  publisher    = {ijcai.org},
  year         = {2024},
  url          = {https://www.ijcai.org/proceedings/2024/586},
  bibsource    = {dblp computer science bibliography, https://dblp.org}
}

@inproceedings{DBLP:conf/iclr/ZhangBHRV17,
  author       = {Chiyuan Zhang and
                  Samy Bengio and
                  Moritz Hardt and
                  Benjamin Recht and
                  Oriol Vinyals},
  title        = {Understanding deep learning requires rethinking generalization},
  booktitle    = {5th International Conference on Learning Representations, {ICLR} 2017,
                  Toulon, France, April 24-26, 2017, Conference Track Proceedings},
  publisher    = {OpenReview.net},
  year         = {2017},
  url          = {https://openreview.net/forum?id=Sy8gdB9xx},
  bibsource    = {dblp computer science bibliography, https://dblp.org}
}

@article{han2018co,
  title={Co-teaching: Robust training of deep neural networks with extremely noisy labels},
  author={Han, Bo and Yao, Quanming and Yu, Xingrui and Niu, Gang and Xu, Miao and Hu, Weihua and Tsang, Ivor and Sugiyama, Masashi},
  journal={Advances in neural information processing systems},
  volume={31},
  year={2018}
}

@inproceedings{younesian2021qactor,
  title={Qactor: Active learning on noisy labels},
  author={Younesian, Taraneh and Zhao, Zilong and Ghiassi, Amirmasoud and Birke, Robert and Chen, Lydia Y},
  booktitle={Asian Conference on Machine Learning},
  pages={548--563},
  year={2021},
  organization={PMLR}
}

@inproceedings{patrini2017making,
  title={Making deep neural networks robust to label noise: A loss correction approach},
  author={Patrini, Giorgio and Rozza, Alessandro and Krishna Menon, Aditya and Nock, Richard and Qu, Lizhen},
  booktitle={Proceedings of the IEEE conference on computer vision and pattern recognition},
  pages={1944--1952},
  year={2017}
}

@inproceedings{xu2025stronger,
  title={Stronger Models are Not Always Stronger Teachers for Instruction Tuning},
  author={Xu, Zhangchen and Jiang, Fengqing and Niu, Luyao and Lin, Bill Yuchen and Poovendran, Radha},
  booktitle={Proceedings of the 2025 Conference of the Nations of the Americas Chapter of the Association for Computational Linguistics: Human Language Technologies (Volume 1: Long Papers)},
  pages={4392--4405},
  year={2025}
}

@article{yao2021instance,
  title={Instance-dependent label-noise learning under a structural causal model},
  author={Yao, Yu and Liu, Tongliang and Gong, Mingming and Han, Bo and Niu, Gang and Zhang, Kun},
  journal={Advances in Neural Information Processing Systems},
  volume={34},
  pages={4409--4420},
  year={2021}
}

@article{xia2020part,
  title={Part-dependent label noise: Towards instance-dependent label noise},
  author={Xia, Xiaobo and Liu, Tongliang and Han, Bo and Wang, Nannan and Gong, Mingming and Liu, Haifeng and Niu, Gang and Tao, Dacheng and Sugiyama, Masashi},
  journal={Advances in neural information processing systems},
  volume={33},
  pages={7597--7610},
  year={2020}
}

@inproceedings{DBLP:conf/iclr/ZhangZW0021,
  author       = {Yikai Zhang and
                  Songzhu Zheng and
                  Pengxiang Wu and
                  Mayank Goswami and
                  Chao Chen},
  title        = {Learning with Feature-Dependent Label Noise: {A} Progressive Approach},
  booktitle    = {9th International Conference on Learning Representations, {ICLR} 2021,
                  Virtual Event, Austria, May 3-7, 2021},
  publisher    = {OpenReview.net},
  year         = {2021},
  url          = {https://openreview.net/forum?id=ZPa2SyGcbwh},
  bibsource    = {dblp computer science bibliography, https://dblp.org}
}

@article{xu2024hallucination,
  title={Hallucination is inevitable: An innate limitation of large language models},
  author={Xu, Ziwei and Jain, Sanjay and Kankanhalli, Mohan},
  journal={arXiv preprint arXiv:2401.11817},
  year={2024}
}

@article{bradley1952rank,
  title={Rank analysis of incomplete block designs: I. the method of paired comparisons},
  author={Bradley, Ralph Allan and Terry, Milton E},
  journal={Biometrika},
  volume={39},
  number={3/4},
  pages={324--345},
  year={1952},
  publisher={JSTOR}
}

@article{liang2022reward,
  title={Reward uncertainty for exploration in preference-based reinforcement learning},
  author={Liang, Xinran and Shu, Katherine and Lee, Kimin and Abbeel, Pieter},
  journal={arXiv preprint arXiv:2205.12401},
  year={2022}
}
